\documentclass[letterpaper]{article} 
\usepackage{aaai23}  
\usepackage{times}  
\usepackage{helvet}  
\usepackage{courier}  
\usepackage[hyphens]{url}  
\usepackage{graphicx} 
\usepackage{natbib}  
\usepackage{caption} 
\usepackage{algorithm}
\usepackage{algorithmic}

\usepackage{newfloat}
\usepackage{listings}
\DeclareCaptionStyle{ruled}{labelfont=normalfont,labelsep=colon,strut=off} 
\floatstyle{ruled}
\newfloat{listing}{tb}{lst}{}
\floatname{listing}{Listing}
\usepackage{amsmath}
\usepackage{amsfonts}
\usepackage{subcaption}
\usepackage{graphicx}
\usepackage{multirow}
\usepackage{booktabs}

\DeclareMathOperator*{\argmax}{arg\,max}

\title{Zero-Shot Assistance in Sequential Decision Problems}
\author {
    Sebastiaan De Peuter,\textsuperscript{\rm 1}
    Samuel Kaski, \textsuperscript{\rm 1, \rm 2}
}
\affiliations {
    \textsuperscript{\rm 1} Department of Computer Science, Aalto University, Espoo, Finland\\
    \textsuperscript{\rm 2} Department of Computer Science, University of Manchester, Manchester, UK\\
    sebastiaan.depeuter@aalto.fi, samuel.kaski@aalto.fi
}

\begin{document}

\maketitle

\begin{abstract}
We consider the problem of creating assistants that can help agents solve new sequential decision problems, assuming the agent is not able to specify the reward function explicitly to the assistant. Instead of acting in place of the agent as in current automation-based approaches, we give the assistant an advisory role and keep the agent in the loop as the main decision maker. The difficulty is that we must account for potential biases of the agent which may cause it to seemingly irrationally reject advice. To do this we introduce a novel formalization of assistance that models these biases, allowing the assistant to infer and adapt to them. We then introduce a new method for planning the assistant's actions which can scale to large decision making problems. We show experimentally that our approach adapts to these agent biases, and results in higher cumulative reward for the agent than automation-based alternatives. Lastly, we show that an approach combining advice and automation outperforms advice alone at the cost of losing some safety guarantees.
\end{abstract}

\section{Introduction}
In this paper we consider the problem of assisting agents in tackling sequential decision problems which they have never encountered before. Human decision makers are routinely faced with this problem. Take for example engineering design~\citep{rao2019}, where one looks to find or construct the best possible design within a space of designs that are feasible. Every design problem is new: each time an architect builds a house it is for different clients. Although the problem is novel to the agent, we can assume that it already knows how to solve it in principle, though not optimally. An architect does not need to re-learn architecture when designing a new house, but can apply their general knowledge and experience to this new design problem.

These design problems can be thought of as single-episode decision problems: they consist of a sequence of decisions, each changing or elaborating a design in some way. Once a satisfactory design has been found, the episode terminates. The decisions are driven by a goal, which can be encoded as a reward function, known only to the agent. This goal is usually tacit and complex, meaning that the agent is unable to provide an accurate explicit description of it.

\begin{figure}[t]
\centering
\includegraphics[width=0.75\columnwidth]{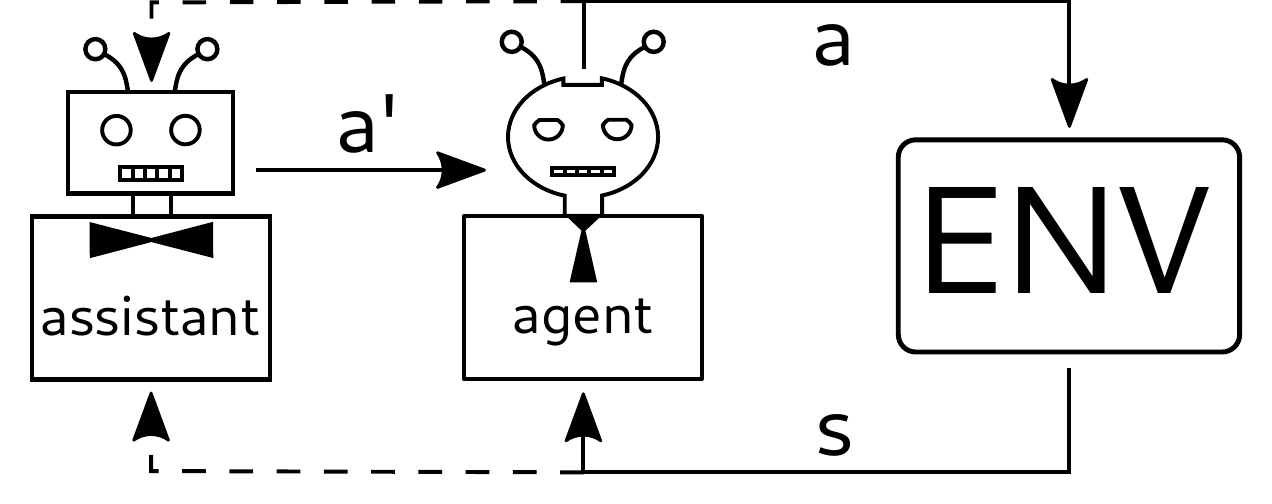}
\caption{In \emph{zero-shot assistance} an assistant helps an agent solve a problem without initially knowing the agent's reward function. We propose an assistant which helps the agent primarily by advising it, leaving the agent in direct control of its environment. In every time step the assistant gives new advice $a'$, appropriate for the current state, based on its inference of the agent's reward function and potential biases. When acting the agent incorporates the advice into its own decision making. The assistant observes both the action $a$ taken by the agent and the new state of the environment $s$, and uses this to infer the agent's reward function and biases.}
\label{fig:interaction}
\end{figure}

We seek to create \emph{assistants} which can assist agents in solving these types of decision problems. Although the assistant is technically an agent, to avoid confusion we will always refer to it as the \emph{assistant} and will only use the term \emph{agent} to refer to the agent being assisted. We think of the agent as an online decision maker who can be assisted in their decision making. The goal for the assistant is to increase the quality of the agent's decisions, measured by cumulative reward, relative to the agent's effort. There are two things the assistant does not know a-priori about the agent: the agent's reward function, and any biases that may cause the agent to deviate from optimal behaviour. Although the agent knows its reward function, it has never solved its problem before -- ruling out inferring the reward function from prior observations -- and is not able to provide an explicit description of the reward function. Similarly, biases are not something an agent is aware of, and thus are not something it can communicate to the assistant. To assist, the assistant must infer both during the episode. Therefore, we call this \emph{zero-shot assistance}. Zero-shot assistance is closely related to zero-shot coordination~\citep{hu2020}, though in the latter the reward function is known to all participants.

We introduce \emph{AI-Advised Decision-making}~(AIAD). In AIAD an assistant helps an agent primarily by giving advice, while the agent remains responsible for taking actions in the environment. Figure~\ref{fig:interaction} shows the interaction between the agent and the assistant. The advice is based on a reward function that is inferred from the agent's behaviour.  For simplicity we will focus on advice of the type "have you considered doing $a$", where $a$ is an action, though AIAD readily generalizes to other forms of advice. We see two fundamental advantages in having an assistant that advises. First, taking advice into account takes little effort while it can be really helpful. Bad advice can be swiftly rejected. Second, it keeps the agent firmly in control and able to reject advice that would have a negative impact. This is a minimum requirement in applications where safety is a concern.

Choosing advice to give to a biased agent requires the assistant to account for those biases. Biases can cause apparently irrational behaviour, including the rejection of useful advice. We consider these biases to include innate limitations or constraints within the agent's decision making process, incorrect problem understanding, or limited knowledge. Incorrect problem understanding, for example, has been shown to cause humans to reject advice which rationally is in their interest~\citep{elmalech2015}. More generally, research in psychology starting in the 1970s has shown that humans exhibit a number of cognitive biases, caused by various heuristics they employ in their decision making, which cause a deviation from optimal behaviour~\cite{kahneman1982,ho2022}. Accounting for specific manifestations of these biases will be especially important when assisting human agents. Prior work on assistance has been able to incorporate agent biases that were known \emph{a priori}~\citep{fern2014,hadfield2016,shah2020}. However, if they are not -- as is the case in zero-shot assistance -- they must be inferred online. To address this we model the uncertainty over biases explicitly, allowing the agent to maintain beliefs over which biases an agent has, and to incorporate present and future beliefs into its planning.

\paragraph{Contributions} In this paper we formalize the assistant's problem of advising agents with unknown reward function and biases as a decision problem. We propose a planning algorithm, a variant of Monte Carlo Tree search~(MCTS), for finding the assistant's policy. To evaluate the practicality of AI-advised decision-making we introduce two decision problems: planning a day trip and managing an inventory of products with stochastic demand. A popular baseline approach for reducing agent effort and improving decision making is to automate, by leaving the decision making entirely to an assistant. When no reward function is available, prior work has proposed to first elicit the reward function~\citep{ng2000,wirth2017}, and then automate. In simulation experiments we show that \textbf{(1)} AIAD significantly outperforms these automation-based baselines. We implement two versions of AIAD: a standard version which only makes recommendations and a hybrid form which has direct access to the decision problem and can therefore automate as well at the cost of agent control. We also show that \textbf{(2)} an assistant which infers and accounts for agent biases outperforms one that does not.

\section{Related work}
\paragraph{Learning reward functions from others}
The main alternative to our proposed approach of advising agents is to take decisions in their place, i.e. automation. For this, the reward function must be known. Thus, before automating one needs to elicit or learn the reward function from the agent. Inverse Reinforcement Learning~(IRL) proposes to learn a reward function directly from observing the agent act~\citep{ng2000,abbeel2004,ramachandran2007,arora2021}. In preference-based elicitation~\citep{wirth2017,christiano2017,brown2019}, the agent is asked which of two trajectories or individual decisions it prefers. The agent is assumed to prefer the trajectory with the highest reward. An alternative is to ask the agent for direct feedback on a single trajectory of decisions~\citep{knox2009,warnell2018}. 
A subset of this literature has looked specifically into the feasibility and utility of learning both agents' reward functions and biases~\citep{evans2015,evans2016}. \citet{chan2021} found that biases can make agents' behaviour more informative of their reward function, and that incorrectly modeling biases can result in poor reward inference. \citet{armstrong2018}, however, show that jointly identifying biases and reward from observations is not always possible. \citet{shah2019} investigate under what assumptions biases can be learnt purely from data.

Where applicable, inference and automation happen in two distinct phases in these works; the elicitation process is not informed by the immediate needs of automation. In an assistance method like ours, both happen at the same time, allowing the assistant to reduce its uncertainty with regards to the agent's biases and reward where it matters for the decisions it needs to make~\citep{shah2020}. We note also that under reward uncertainty, an automating policy must necessarily be more risk-averse than an assistant that gives advice, as the automating policy cannot rely on the agent to prevent it from making bad decisions.
\paragraph{Human-AI collaboration}
Our work fits within a larger body of approaches which consider collaboration between an assistant and an agent to solve a common problem. \citet{dimitrikakis2017} consider a setting in which an assistant acts autonomously but can be overridden by an agent at a cost. \citet{celikok2022} consider a similar problem in partially observable environments. Both, however, assume the assistant already knows the reward function. Others have considered collaboration when the assistant does not know the agent's reward function. Shared autonomy~\citep{javdani2015,reddy2018} considers a setting in which the agent gives commands to the assistant, which then acts in the environment. As the assistant does not necessarily follow the commands directly, but uses them to infer the agent's reward function which it then maximizes, we consider this an automating approach. In Cooperative Inverse Reinforcement Learning~\citep{hadfield2016} an assistant and agent jointly solve a problem. The assistant uses IRL to learn the agent's reward function. \citet{fern2014} has proposed assistants which assist agents (not necessarily through advice) in decision problems; \citet{shah2020} proposed similar assistants for partially observable settings. These last three works are similar to ours but assume that, except for the reward function, everything that determines the agent's policy -- including any potential biases --  is known \emph{a priori}. Unlike our method, these methods not support the online inference of biases needed for zero-shot assistance.

\section{Problem setup}
We consider an agent solving a decision problem which can be modeled as an infinite-horizon MDP $E = \langle \mathcal{S}, \mathcal{A}, T, \mathcal{R}_\omega, \gamma, p_{0,s} \rangle$. Here $\mathcal{S}$ is a set of states and $\mathcal{A}$ is the set of actions available to the agent. At time step $t$ the transition function $T(s_{t+1} \mid s_t, a_t)$ defines a distribution of potential next states $s_{t+1}$ given that the agent has taken action $a_t$ in current state $s_t$. $\mathcal{R}_\omega(s_t, a_t, s_{t+1})$ is the reward function. It defines the instantaneous reward for taking action $a_t$ in state $s_t$ and ending up in $s_{t+1}$. $\gamma \in (0,1]$ is the discounting rate. The agent's objective is to maximize its expected discounted cumulative reward $\mathbb{E} \left [ \sum_{t=0}^\infty r_t \gamma^t \mid T, p_{0,s} \right ]$ where $r_t$ is the reward it achieved at time step $t$. Finally, $p_{0,s}$ is the start state distribution: $s_0 \sim p_{0,s}$.

When assisting an agent we will assume that we know certain things about that agent's problem $E$. In line with prior work~\citep{abbeel2004} we assume that we know $\mathcal{S}$, $\mathcal{A}$, $T$, $\gamma$ and $p_{0,s}$. Though we do not know $\mathcal{R}_\omega$, we do have access to its parametric function class $\mathcal{R} = \{\mathcal{R}_{\omega'}\}_{\omega' \in \Omega}$. Note that this assumption is not particularly restrictive; $\mathcal{R}$ could be the space of all reward functions.

\section{AI-advised decision-making (AIAD)}
We now formalize AI-Advised Decision-making. As shown in Figure~\ref{fig:interaction}, the agent acts in $E$ based on advice from the assistant. The goal of the assistant is to maximize the cumulative discounted reward obtained by the agent through this advice. We define this from the assistant's point of view as a decision problem with reward function $\mathcal{R}_\omega(s_t, a_t, s_{t+1})$ and as actions the advice it can give.

For the assistant to be able to plan, we will assume we have an \emph{agent model} ${\hat{\pi}(a \mid s, a'; \theta, \omega)}$ available, a model of the agent's fixed policy upon receiving advice $a'$. This could be an expert-created model based on appropriate assumptions, or could have been learned based on observed agent behaviour on similar problems. It depends on two sets of unobserved parameters: $\omega \in \Omega$ and $\theta \in \Theta$. We defined $\Omega$ earlier as the parameter space of the reward function. $\Theta$ is the parameter space for the possible biases the agent has. We call $\theta$ the bias parameters.

\subsection{Advice as a decision problem}
We define the assistant's decision problem as a generalized hidden parameter MDP~(GHP-MDP)~$\mathcal{M}$~\citep{perez2020}. A GHP-MDP is an MDP in which both the transition and reward function are parameterized but where the true values of those parameters are not observed. In our definition these parameters are $\omega$ and $\theta$, the two unobserved parameters which determine an agent's reward function and biases. We define $\mathcal{M}$ such that the instance $\mathcal{M}_{\omega, \theta}$ defines the problem of assisting an agent with reward parameters $\omega$ and bias parameters $\theta$.

Let $\mathcal{M} = \langle \mathcal{S}, \Omega, \Theta, \mathcal{A}', \mathcal{A}, \mathcal{T}, \hat{\pi}, \mathcal{R}, \gamma, p_{0,s}, p_{0,\omega}, p_{0,\theta} \rangle$. Here $\mathcal{S}, \mathcal{A}, \gamma$, and $p_{0,s}$ are the same as in the agent's problem $E$. $\mathcal{R}$ and $\hat{\pi}(a \mid s, a'; \theta, \omega)$ are as defined earlier. The assistant's actions $\mathcal{A}'$ constitute advice that can be given to the agent. $\Omega$ and $\Theta$ are the parameter spaces for the reward function and biases, and $p_{0,\omega}$ and $p_{0,\theta}$ are prior distributions over them. Lastly, $\mathcal{T}$ is a collection of transition functions $\mathcal{T}_{\omega,\theta}$ for all possible values of $\omega \in \Omega$ and $\theta \in \Theta$. It encodes the interaction between the assistant and the agent and between the agent and the environment from Figure~\ref{fig:interaction}.

When the assistant gives advice $a_t' \in \mathcal{A'}$ to the agent, the agent is free to choose which action $a_t \in \mathcal{A}$ to take in $E$. It is this action taken by the agent that leads to a new state, according to the transition function of $E$. Thus, the assistant only indirectly influences the change of state, by using advice to induce a different policy from the agent. The agent model $\hat{\pi}$ predicts which policy will be induced by advice. Thus, for given reward and bias parameters $\omega$ and $\theta$, the transition function is \[\mathcal{T}_{\omega, \theta}(s_{t+1} \mid s_t, a_t') = \!\! \sum_{a_t \in \mathcal{A}} \! \hat{\pi}(a_t \mid s_t, a_t'; \theta, \omega) T(s_{t+1} \mid s_t, a_t)\]

Because the true values of $\theta$ and $\omega$ are not known, we can think of $\mathcal{M}$ as defining a space of MDPs. The challenge in planning over $\mathcal{M}$ is that planning must happen without knowing which MDP in this space the assistant is truly operating in, i.e. what kind of agent the assistant is advising. We can, however, maintain beliefs over $\theta$ and $\omega$ based on the transitions we observe. For every transition $(s_t, a_t', s_{t+1})$ we observe we can calculate the likelihood of that transition under the various possible parameter values in $\Omega \times \Theta$ to update our posterior belief distributions over the parameters. In other words, by observing the agent's decisions in response to advice we can maintain beliefs about the agent's biases and reward function.

\subsection{Root sampling for GHP-MDPs}
Finding an optimal policy over $\mathcal{M}$ involves not only planning on a belief distribution over MDPs, but also accounting for how that distribution will change as we act. With every action we take, we observe a new transition which will change our beliefs over $\Omega$ and $\Theta$. However, not every action is equally informative: advice which gets wildly different reactions from different types of agents will be more useful for determining what type of agent we are assisting than advice to which all agents react in the same way. Planning must consider both the expected long-term reward of actions, and their informativeness towards the unknown parameters.

We propose a modification of \emph{Bayes-adaptive Monte Carlo Planning} (BAMCP)~\citep{guez2012}. BAMCP is based on MCTS~\citep{browne2012} and enables planning over MDPs where the transition function is not known, and needs to be inferred from transition observations. The advantage of BAMCP is that it efficiently maintains all current and potential future beliefs over transition functions by using the tree as a particle filter~\citep{guez2012}. This allows it to incorporate the future information value of actions into its value estimates. We extend this algorithm from operating on beliefs over transition functions to joint beliefs over transition and reward functions, i.e. beliefs over $\Theta$ and $\Omega$. We call this new variant \emph{Generalized Hidden Parameter Monte Carlo Planning} (GHPMCP).

We give a short overview of the algorithm here, and refer the reader to the appendices for a detailed explanation. Like any MCTS algorithm, in every planning iteration GHPMCP simulates an MDP down the tree following a UCT policy. The main difference is that the simulated MDP is resampled for every iteration. Before an iteration starts, parameters $\theta,\omega$ are sampled from $p_\theta, p_\omega$. $\mathcal{M}_{\omega,\theta}$ is then simulated down the tree. The Q-function estimates along the path are updated using $\mathcal{M}_{\omega,\theta}$'s specific reward function $R_\omega$. Here lies the difference to BAMCP, which -- because it only considers uncertainty over the transition function -- uses the same fixed reward function $R$ in every iteration.

\section{An agent model for assistance}
We now introduce a general-purpose agent model, applicable to any decision problem $E$ as defined above. We have developed this agent model to be a good starting point for most use cases. It is based on established and grounded theories of human decision making. It is also consistent with how RL agent policies are often implemented. We use instances of it in our experiments here, but stress that our proposed method does not require this agent model specifically.

The proposed agent model is built on the following choice rule:
\begin{equation}
    \label{eq:boltzmann_rationality}
    p(a | u) = \frac{p(a) \exp \left(\beta u(a)\right)}{\sum_{\hat{a} \in \mathcal{A}} p(\hat{a}) \exp \left(\beta u(\hat{a})\right)}
\end{equation}
where $a \in \mathcal{A}$ is an action, $u$ is a function that assigns a utility to every action, $p(a)$ is a prior distribution over actions, and $\beta$ is a temperature parameter. $\beta$ allows us to interpolate between fully rational choices ($\beta = \infty$) and fully random choices ($\beta = 0$).  This choice rule has repeatedly proven to be a useful and practical model of human cognition~\citep{lucas2008,baker2009,viappiani2010,christiano2017,brown2019}. Theoretical work has proposed it as a result of a bounded rational view of human cognition with information processing costs~\citep{ortega2013,genewein2015}. Surprisingly, this rule is also a popular choice of model in RL -- where it is known as a softmax policy -- for representing agent policies~\citep{sutton2018}.

The agent model ${\hat{\pi}(a \mid s, a'; \theta, \omega)}$ consists of a sequence of choices made according to this choice rule. It is based on the idea that an agent will only settle for the assistant's action if it cannot find a better one itself. As utility function we use the Q-function of the current state $u(a) = \hat{Q}(s, a; \theta, \omega)$. This Q-function could be derived directly from the agent's problem $E$, or from some derivative of it $\hat{E}$ in case we want to model an agent with an incorrect or limited view of the world. Some biases can therefore be modeled as part of $\hat{E}$. $\hat{Q}$ then depends on $\omega$ to allow us to model different reward functions, and on $\theta$ so that we can use the bias parameters as parameters of the problem model $\hat{E}$.

Under our model the agent starts by choosing the best action it can think of. This is a stochastic choice which we represent as a random variable $A_1$ defined over the actions. The distribution of $A_1$ results from a straightforward application of equation~\eqref{eq:boltzmann_rationality}:
\begin{equation}
    \label{eq:own_action_choice}
    p_{A_1} \left (a \right ) = \frac{p(a) \exp \left ( \beta_1 \hat{Q}(a) \right )}{\sum_{\hat{a} \in \mathcal{A}}p(\hat{a}) \exp \left ( \beta_1 \hat{Q}(\hat{a}) \right )}
\end{equation}
As the state and parameters are constant within this context, we have used the shorthand $\hat{Q}(a) := \hat{Q}(s, a; \theta, \omega)$ and $p(A_1 = a) := p(A_1 = a | s; \theta, \omega)$. We will drop the conditioning variables $s, \theta$ and $\omega$ for the rest of this section.

Next the agent chooses whether to switch to the assistant's recommended action $a'$ or to stick to the action $a$ it has chosen. The probability of switching from $a$ to $a'$ (denoted $a \to a'$) is
\begin{equation}
    p \left ( a \to a' \right ) = \frac{\exp \left ( \beta_2 \left ( \hat{Q}(a') - \hat{Q}(a) \right ) \right )}{1 + \exp \left ( \beta_2 \left ( \hat{Q}(a') - \hat{Q}(a) \right ) \right )} \label{eq:assistance_choice}
\end{equation}
This probability is the result of applying equation~\eqref{eq:boltzmann_rationality} to the binary choice of switching or not with a uniform prior. The utility for both choices is the gain in Q-value realised: $\hat{Q}(a') - \hat{Q}(a)$ in the case of switching and $0$ otherwise. As this choice is easier than the choice in equation~\eqref{eq:own_action_choice} we use a different temperature parameter $\beta_2$ here. We represent the agent's choice of action after considering $a'$ by $A_2$, a random variable with distribution
\[p_{A_2}( a | a') = \begin{cases}
    [1-p(a \to a')] p_{A_1}(a) & \text{if } a \neq a' \\
    [1-p(a' \to a')] p_{A_1}(a')\\
    \;\;\, + \sum_{a'' \in \mathcal{A}} p(a'' \to a') p_{A_1}(a'') & \text{if } a = a'\\
  \end{cases}\]
This distribution is then the agent model's policy $\hat{\pi}$.

In some problems, such as design problems, there is a special NOOP action which allows the agent to choose to do nothing. We model this as the agent recommending the NOOP action to itself, after arriving at $A_2$, analogously to the switch to the assistant's recommendation $a'$.

\section{Experiments}
We present here results from a number of simulation experiments on two single-episode decision problems: a day trip design problem and an inventory management problem\footnote{Our code is available from \url{https://github.com/AaltoPML/Zero-Shot-Assistance-in-Sequential-Decision-Problems}} Both exhibit a very different structure. The day trip design problem has vast state and action spaces that represent a design that evolves over time. Actions do not have an inherent cost, but rather the value of the design produced at the end of the episode is important. Returning to the start state (the starting design) without restarting the episode is trivial. The inventory management problem on the other hand has a smaller state space but all actions contribute to the cumulative reward of the episode, and it is generally impossible to reset the problem back to its starting point within an episode.

We will compare agents assisted by AIAD to agents assisted by a number of baselines on these two decision problems. We consider two versions of AIAD: \textbf{AIAD} and \textbf{AIAD + automation}. The latter is an extension which gives the assistant actions that directly change the environment. These actions have transition function $\mathcal{T}_{\omega, \theta}(s_{t+1} \mid s_t, a_t') = T(s_{t+1} \mid s_t, a_t')$. We consider four baselines: \textbf{(1) unassisted agent} To create an unassisted agent we modify our agent model by removing the switch to a recommended action encoded in eq.~\eqref{eq:assistance_choice} ($p(a \to a') = 0 \; \forall a \in \mathcal{A}$). \textbf{(2) IRL + automation}. This is an IRL-based approach following prior work in learning rewards from biased agents. It observes $N$ time steps of the unassisted agent acting without assistance. It then infers both $\theta$ and $\omega$ from the observations, using the same agent model but without knowledge of the parameters, and completes the rest of the episode in place of the agent. For day trip design we first return back to the starting design before automating. The automation policy is an optimal policy for the agent's problem $E$ with as reward function the expected reward under the inferred posterior over reward parameters. \textbf{(3) PL + automation}. This approach is based on preference learning~(PL). At the start of the episode the agent is presented with $N$ comparison queries. These queries are selected based on their expected information gain. We create an agent model that chooses between the two options according to the choice rule from eq.~\eqref{eq:boltzmann_rationality} with $u(s) = f_\omega(s)$. This model is both used to simulate the agent and to infer the reward function from the agent's responses. We then automate all decisions within the episode based on the inferred reward function. \textbf{(4) partial automation} This method is a more flexible version of IRL + automation. It automates by default, but in any time step can temporarily hand control back to the agent. The agent then acts -- without advice -- in that time step. This allows partial automation to rely on the agent when it is too uncertain about how to act. The agent's observed action is used to update the beliefs about $\theta$ and $\omega$. This baseline is representative for approaches such as \cite{shah2020,hadfield2016}, albeit re-implemented here within our framework so that biases can be inferred.

Every run of our experiments lasts for one episode. Posterior beliefs in all implementations are maintained using a weighted particle filter. Within a run every method is applied once to assist a simulated agent in an instance of the problem considered. The cumulative reward obtained through different assistance methods is then compared using a paired Wilcoxon signed rank test at significance level $p < 0.05$.

\subsection{Day Trip Design}
The day trip design problem is an idealized but otherwise realistic instance of a design problem. Like most other design problems this has a large action space and a vast state space. The agent is given 100 points of interest (POI) and must choose a subset of them which it wants to visit within a day. Its goal is to choose a subset that it would maximally enjoy visiting. Every POI has a location, visit duration, admission cost, and belongs to a number of topics. There are 20 topics in total. The enjoyment of visiting a POI is a function of the overlap between the topics to which it belongs and the topics the agent is interested in. The total enjoyment of visiting the POIs must be traded off against the sum of their admission costs. The value of a trip $s$, $f_\omega(s)$, is therefore a combination of these two scores, parameterized by parameters $\omega$ which capture the agent's topic interests and tolerance for high admission prices. Choosing POIs involves accounting for the time needed to travel between the chosen points. Any time spend walking cannot be spend enjoying a visit to a POI. To help with this, the agent is automatically given an optimal itinerary which minimizes the travel time for its current selection of POIs (this involves solving a traveling salesperson problem~(TSP)).

To formalize this as an MDP $E$ we define the state space as the space of all subsets of the POIs. We introduce an action for every POI which allows the agent to add that POI to the current trip or to remove it, depending on whether it is already part of it or not. To enforce the constraint that all selected POIs must be visited within a day, in states corresponding to a trip that would take more than 12 hours we only allow actions that remove POIs. The agent seeks to maximize $f_\omega(s)$. Therefore, we define its reward function as the improvement in objective value from one time step to the next: $R_\omega(s_t, a_t, s_{t+1}) = f_\omega(s_{t+1}) - f_\omega(s_t)$. Additional details about this experiment can be found in the appendices.

We use the agent model we introduced in the previous section. The problem $\hat{E}$ on which the agent plans differs in two key aspect from $E$. The first is that the agent uses a visual heuristic to determine how the itinerary for its current choice of POIs will change as it considers additions and deletions. We model this using a visual heuristic for TSPs commonly used by humans~\citep{macgregor2000}. The second difference is that we introduce an anchoring bias. This bias, which is typical in human designers, causes the agent to resist large changes to its design. In our implementation, agents with this bias will refuse to consider adding any POI that is more than 500 meters away from their current itinerary. Concretely, this means that in a state $s$ in $\hat{E}$, the available actions will only be those that add POIs which are within 500 meters of the itinerary, or that remove POIs. The Q-values used in the agent model are determined using depth-limited best-first search on $\hat{E}$.

Because this is a design problem, the cost of taking actions -- i.e. changing the design -- is non-existent. We are therefore mainly interested in minimizing the effort required from the agent; how many actions the assistant takes is not a significant factor. Thus, our quantity of interest is the objective value $f_\omega(s)$ achieved as a function of the number of agent interactions $N$. Due to our definition of the reward function, this is equal here to the undiscounted cumulative reward after $N$ interactions. As interactions we count both actions taken and queries answered by the agent. For PL, queries consist of of a comparisons between two day trips (i.e. states). We have created a separate agent model that chooses between the two day trips according to the choice rule from eq.~\eqref{eq:boltzmann_rationality} with $u(s) = f_\omega(s)$. The PL and IRL baselines are evaluated at 0, 5, 10 15, 20, 25 and 30 interactions, while the other methods are evaluated on a continuous range of $N$ from 1 to 30. We ran this experiment 75 times. For every run we sampled new agent model parameters $\theta,\omega$ and a new set of POIs. Roughly half of the agents simulated in these runs had an anchoring bias.

\subsubsection{Results}
\begin{figure*}[t]
    \centering
    \begin{subfigure}[b]{0.43\textwidth}
        \includegraphics[width=\textwidth]{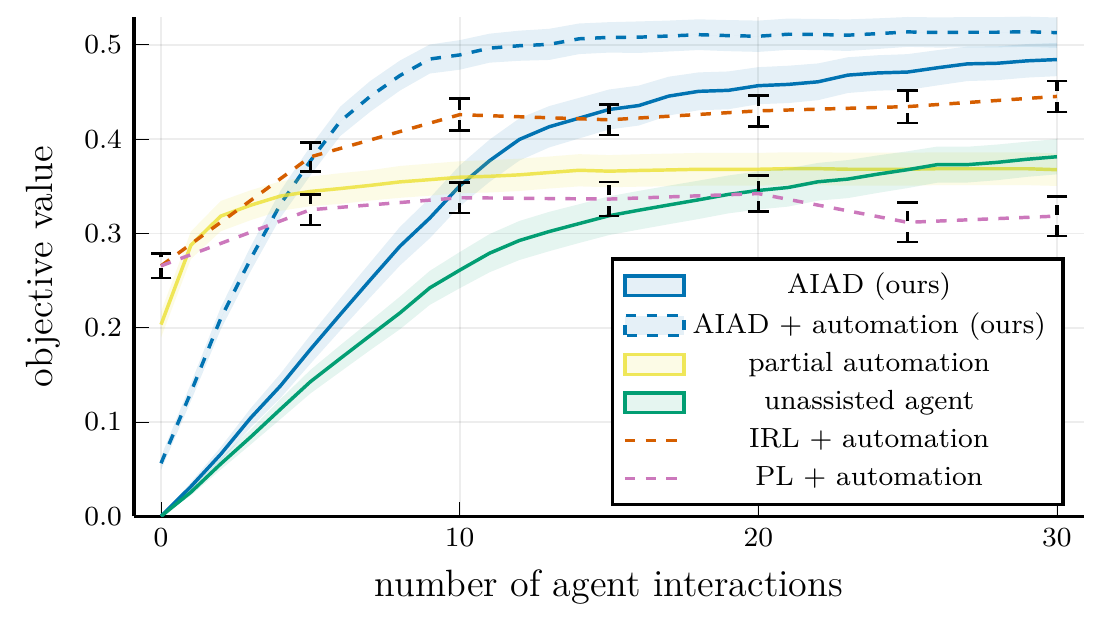}
        \caption{}
        \label{fig:anchoring_experiment_performance}
    \end{subfigure}
    ~
    \begin{subfigure}[b]{0.43\textwidth}
        \includegraphics[width=\textwidth]{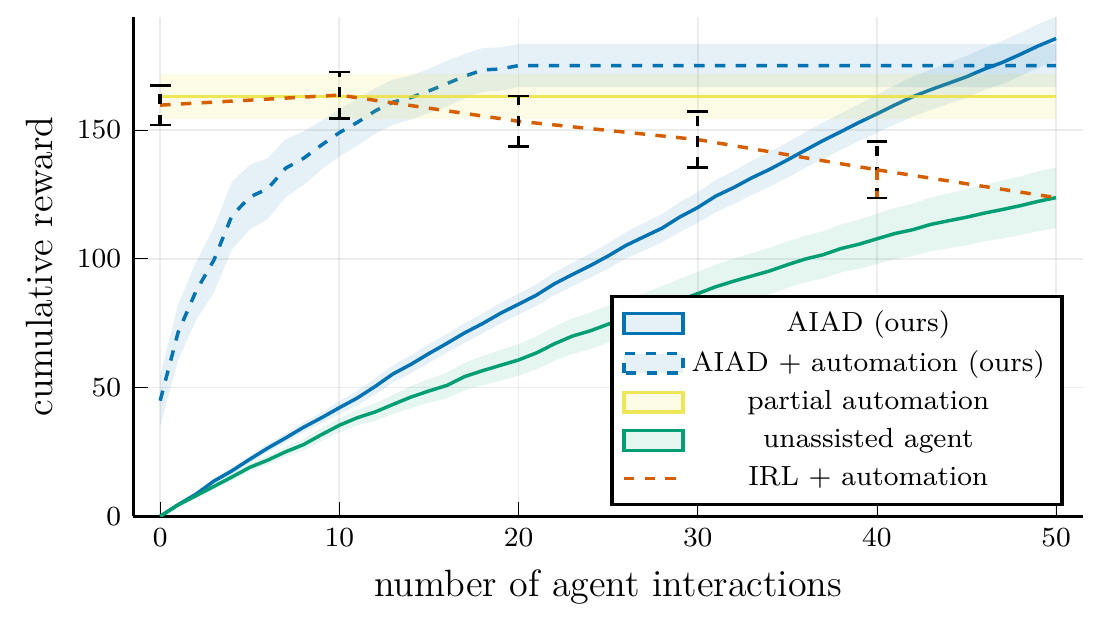}
        \caption{}
        \label{fig:inventory_management_cum_rew_per_interaction}
    \end{subfigure}
    \caption{\textbf{(a)} Mean objective value achieved by agents supported by the methods considered as a function of different numbers of interactions for the day trip design problem. This plot only shows agent interactions. Changes in objective value achieved by the assistant are added to the last agent action that preceded them. The Shading shows the standard error around the mean. \textbf{(b)} Cumulative discounted reward achieved by agents supported by the methods considered as a function of the number of agent interactions for the inventory management problem.}
\end{figure*}
We observe that agents assisted by AIAD achieved significantly higher objective value than those assisted by any of the baselines from 18 interactions onward (figure~\ref{fig:anchoring_experiment_performance}). Because every episode starts from an empty trip, and AIAD needs to interact with the agent to change the design, there is a minimum number of agent interactions required to produce a complete design. This puts it at a disadvantage to the automation-based baselines which can change the design directly. AIAD + automation addresses this by allowing the assistant to change the design directly. This also allows the assistant to bypass the agent when the anchoring bias would prevent the agent from accepting certain recommendations. We see that AIAD + automation significantly outperforms all baseline-assisted agents from 8 interactions. Although these results are a clear improvement over AIAD, AIAD + automation lacks the safety guarantees and agent control of AIAD, as the agent cannot interrupt the assistant if it automates poorly.

Though AIAD achieves lower uncertainty with regards to the agent parameters and lower mean loss in its inference of the reward (see additional results in the appendices), compared to the automation-based baselines, we consider these differences too small to explain its performance edge. We see a larger part of the explanation in a specific advantage of assistance through advice. Bad advice -- as a result of uncertain inferences of the parameters or mis-specification of the agent model -- does not usually affect the design, as the agent should simply ignore it, whereas bad direct changes to the design will. This is true for AIAD + automation too, which can use its uncertainty estimates to decide whether to use advice or make a direct change. In our experiments the reward function was technically mis-specified, as the particles in the particle filter could not cover the whole reward parameter space. We see evidence of this for all methods in the low uncertainty and high inference error in the latter part of the episode.

We analyze in the appendices the relation between the loss in the inferred reward function and the rejection rate of advice, and show that the rejection rate decreases as the inference of the reward function in AIAD improves. Of course, this inherent safety of advice works only if the agent is likely to reject bad advice. This was the case in the experiments considered here, but in other instances specific biases may cause an agent to follow such bad advice. Luckily, the assistant can account for this if these biases are modeled in the agent model.

To verify that an assistant that infers and accounts for the anchoring biases is more helpful than one that does not we ablate our AIAD implementation by considering two alternatives which make assumptions about this bias, rather than inferring it. One assumes that no agents have an anchoring bias and does not model it, and the other assumes that all have it. Both were significantly outperformed by standard AIAD after 11 interactions. More details can be found in the appendices.

\subsection{Inventory Management}
In the inventory management problem the agent is tasked with managing an inventory of three different products. In every time step $t$ every product $i$ has some amount of demand $d_{i,t}$ sampled from a known demand distribution $D_{i,t}$. At the start of every time step, before observing $d_{i,t}$, the agent must decide how much of each product to produce, so that the the current inventory of product $i$, $I_{i,t}$, and the chosen production quantity $P_{i,t}$ are sufficient to meet the demand. There is however a limit on how much product can be produced in total within any given time step. Every piece of product that is sold yields a profit $v_i$, and any unsold product goes into the inventory. Storing a piece of product has a cost $c$ per time step. Any demand that cannot be met from inventory and production on a given day is lost, and the agent incurs a future loss of business cost $l$. Imagine if the agent is running a bakery and sells out of croissants before the end of the day. Any customer who had wanted to buy croissants will have to go to a competitor, and thus may not come back. The future loss of business cost represents these lost sales. 

This problem can be defined as an MDP $E$ with as state the current inventory for all products $\{I_{i,t}\}_{i = 1:3}$ and all future demand distributions $\{D_{i,t}\}_{i=1:3,t}$. The actions represent choices of how much of each product to produce. In our implementation case the agent is able to produce any multiple of 2 of any product, up to a sum of 12. The reward function is parameterized by $c, l$ and $\{v_i\}_{i=1:3}$.

We use the agent model we introduced in the previous section. The problem $\hat{E}$ on which the agent plans differs from $E$ in that we assume the agent has insufficient computational resources to work with the full demand distributions $D_{i,t}$, and instead plans using point estimates of the demand $\hat{d}_{i,t}$. We define $\hat{d}_{i,t} = \mu(D_{i,t}) + \theta \sigma(D_{i,t})$ where $\mu$ and $\sigma$ are respectively the mean and standard deviation of $D_{i,t}$. $\theta \in \Theta$ is a single continuous bias parameter. When $\theta \neq 0$ it introduces a bias into the agent model, specifically an optimism bias ($\theta > 0$) under which the agent overestimates the expected demand, or a pessimism bias ($\theta < 0$) under which the agent underestimates the demand. These biases are typical for humans. Optimism could lead to excessive production, and therefore excessive cost of storing unsold inventory, while pessimism could cause the agent to incur a high future loss of business cost due to insufficient production. The Q-values used in this agent model are determined using depth-limited best-first search on $\hat{E}$.

Our main quantity of interest for this problem is the discounted cumulative reward, both calculate over the whole episode, and as a function of the number of agent actions. We ran this experiment 20 times, with episodes of 50 time steps. For the IRL baseline we change from agent to automation after 0, 10 20, 30, and 40 actions taken by the agent. For every run we sample new parameters $\theta,\omega$ and new demand distributions. The optimism/pessimism bias parameter $\theta$ is sampled from a zero-centered Gaussian distribution. Additional details and results can be found in the appendices.

\subsubsection{Results}
\begin{table}[ht]
\centering
\begin{tabular}{r c}
method & cumulative reward\\
\midrule
AIAD & $\boldsymbol{185.5 \pm 8.5}$ \\
AIAD + automation & $175.0 \pm 8.3$ \\
unassisted agent & $123.7 \pm 11.7$ \\
IRL + automation & $165.6 \pm 9.0$ \\
partial automation & $163.1 \pm 8.7$ \\
\midrule
oracle + automation & $187.6 \pm 7.6$ \\
\end{tabular}
\caption{Mean cumulative discounted reward ($\pm$ standard error) achieved by agents supported by the methods considered on the inventory management problem over a complete episode. Bold indicates that the method is significantly better than the baselines. For IRL we show the best achieved result, which switched to automation after 10 interactions.}
\label{tab:inventory_management_cumulative_reward}
\end{table}
Table~\ref{tab:inventory_management_cumulative_reward} shows the mean cumulative discounted reward for episodes of the inventory management problem. We can see that if we are not trying to minimize agent effort and only aim to maximize cumulative discounted reward, AIAD significantly outperforms all other methods. In fact, AIAD comes very close to automation based on the true reward function (\textbf{oracle + automation}). Looking at cumulative discounted reward as a function of the number of agent interactions (figure~\ref{fig:inventory_management_cum_rew_per_interaction}) the picture is different. For low levels of agent effort it is best to automate based on the prior, i.e. without interacting with the agent at all (this is represented by IRL + automation at 0 interactions). From 19 interactions onward AIAD + automation significantly outperforms all the other methods, until 50 interactions where standard AIAD significantly outperforms it.

To verify that an assistant that infers and accounts for an optimism/pessimism bias is more helpful than one that does not we compare AIAD to three ablations which assume a certain bias. The first assumes that all agents are optimistic, the second that all agents are pessimistic and the last that no agents are biased. We find that agents assisted by AIAD achieve significantly higher cumulative reward if AIAD infers this bias. More detail can be found in the appendices.

\section{Conclusion}
In this paper we have considered zero-shot assistance: the problem of assisting an agent in a decision problem when no prior knowledge of the agent's reward function or biases is available. To this end we have introduced \emph{AI-Advised Decision-making}~(AIAD), in which an assistant helps an agent primarily by giving advice. We also introduced a version of AIAD which allowed the agent to automate, at the cost of losing some of the safety guarantees of AIAD. We have introduced a decision-theoretic formalization of the assistant's problem of advising such an agent, and have proposed a planning algorithm for determining the assistant's policy. An important novelty in this formalization is that it accounts for individual agent biases, something which we showed experimentally improves the quality of the assistant's advice. Through our experiments we have also shown that assistance through advice, potentially combined with some automation, yields better results than assistance through automation alone.

\paragraph{Limitations} Although our work does not require the explicit definition of a reward function, we do require the definition of a space of reward functions, which may still be difficult to provide. This difficulty is, however, inherent to any reward learning approach. Though our framework supports advice of any type, our experiments only covered action recommendations. Other types of advice could be designed to push an agent's reasoning in a general direction rather than toward a single action, or could include additional information (such as visualizations) designed to convince the agent of the quality of an action recommendation. We leave this to future work. Our experiments did not cover the effects of misspecification in the agent model itself, only in the reward function. For complex agent like humans, it is unlikely that we would be able to create a perfect agent model.

\section{Ethical Statement}
Assistance through advice is a promising solution for the value alignment problem~\citep{everitt2018}. As we have discussed in this paper, advice can reduce the negative effects of value misalignment in an assistant while ensuring agent (human) control. Further, advice forces the assistant to be understandable, as it must convince the agent to follow its advice. Equipping an assistants with a highly accurate agent model -- whether in AIAD or other methods -- does pose some safety risks. There is the potential for an assistant to use this model to find ways to weaken the agent's control, for example by exploiting its biases.

\section{Acknowledgements}
We would like to thank Pierre-Alexandre Murena and Mustafa Mert \c{C}elikok for their valuable feedback and suggestions. This works was supported by the Technology Industries of Finland Centennial Foundation and the Jane and Aatos Erkko Foundation under project Interactive Artificial Intelligence for Driving R\&D, the Academy of Finland (flagship programme: Finnish Center for Artificial Intelligence, FCAI; grants 328400, 345604 and 341763), and the UKRI Turing AI World-Leading Researcher Fellowship, EP/W002973/1. We further acknowledge the computational resources provided by the Aalto Science-IT project.

\bibliography{references}

\begin{thebibliography}{38}
\providecommand{\natexlab}[1]{#1}

\bibitem[{Abbeel and Ng(2004)}]{abbeel2004}
Abbeel, P.; and Ng, A.~Y. 2004.
\newblock Apprenticeship {L}earning via {I}nverse {R}einforcement {L}earning.
\newblock In \emph{Proceedings of the Twenty-First International Conference on
  Machine Learning}.

\bibitem[{Armstrong and Mindermann(2018)}]{armstrong2018}
Armstrong, S.; and Mindermann, S. 2018.
\newblock Occam's razor is insufficient to infer the preferences of irrational
  agents.
\newblock \emph{Advances in Neural Information Processing Systems}, 31.

\bibitem[{Arora and Doshi(2021)}]{arora2021}
Arora, S.; and Doshi, P. 2021.
\newblock A survey of inverse reinforcement learning: {C}hallenges, methods and
  progress.
\newblock \emph{Artificial Intelligence}, 297.

\bibitem[{Baker, Saxe, and Tenenbaum(2009)}]{baker2009}
Baker, C.~L.; Saxe, R.; and Tenenbaum, J.~B. 2009.
\newblock Action understanding as inverse planning.
\newblock \emph{Cognition}, 113(3): 329--349.

\bibitem[{Brown et~al.(2019)Brown, Goo, Nagarajan, and Niekum}]{brown2019}
Brown, D.; Goo, W.; Nagarajan, P.; and Niekum, S. 2019.
\newblock Extrapolating {B}eyond {S}uboptimal {D}emonstrations via {I}nverse
  {R}einforcement {L}earning from {O}bservations.
\newblock In \emph{Proceedings of the 36th International Conference on Machine
  Learning}, 783--792. PMLR.

\bibitem[{Browne et~al.(2012)Browne, Powley, Whitehouse, Lucas, Cowling,
  Rohlfshagen, Tavener, Perez, Samothrakis, and Colton}]{browne2012}
Browne, C.~B.; Powley, E.; Whitehouse, D.; Lucas, S.~M.; Cowling, P.~I.;
  Rohlfshagen, P.; Tavener, S.; Perez, D.; Samothrakis, S.; and Colton, S.
  2012.
\newblock A {S}urvey of {M}onte {C}arlo {T}ree {S}earch {M}ethods.
\newblock \emph{IEEE Transactions on Computational Intelligence and AI in
  games}, 4(1): 1--43.

\bibitem[{{\c{C}}elikok, Oliehoek, and Kaski(2022)}]{celikok2022}
{\c{C}}elikok, M.~M.; Oliehoek, F.~A.; and Kaski, S. 2022.
\newblock Best-{R}esponse {B}ayesian {R}einforcement {L}earning with
  {B}ayes-adaptive {POMDPs} for Centaurs.
\newblock In \emph{Proceedings of the 21st International Conference on
  Autonomous Agents and Multiagent Systems}, 235--243.

\bibitem[{Chan, Critch, and Dragan(2021)}]{chan2021}
Chan, L.; Critch, A.; and Dragan, A. 2021.
\newblock Human irrationality: both bad and good for reward inference.
\newblock \emph{arXiv preprint arXiv:2111.06956}.

\bibitem[{Christiano et~al.(2017)Christiano, Leike, Brown, Martic, Legg, and
  Amodei}]{christiano2017}
Christiano, P.~F.; Leike, J.; Brown, T.~B.; Martic, M.; Legg, S.; and Amodei,
  D. 2017.
\newblock Deep {R}einforcement {L}earning from {H}uman {P}references.
\newblock In \emph{Proceedings of the 31st International Conference on Neural
  Information Processing Systems}, 4302--4310.

\bibitem[{Dimitrakakis et~al.(2017)Dimitrakakis, Parkes, Radanovic, and
  Tylkin}]{dimitrikakis2017}
Dimitrakakis, C.; Parkes, D.~C.; Radanovic, G.; and Tylkin, P. 2017.
\newblock Multi-{V}iew {D}ecision {P}rocesses: {T}he {H}elper-{AI} {P}roblem.
\newblock In \emph{Advances in Neural Information Processing Systems},
  volume~30.

\bibitem[{Elmalech et~al.(2015)Elmalech, Sarne, Rosenfeld, and
  Erez}]{elmalech2015}
Elmalech, A.; Sarne, D.; Rosenfeld, A.; and Erez, E.~S. 2015.
\newblock When {S}uboptimal {R}ules.
\newblock In \emph{Proceedings of the Twenty-Ninth AAAI Conference on
  Artificial Intelligence}, 1313--1319. AAAI.

\bibitem[{Evans and Goodman(2015)}]{evans2015}
Evans, O.; and Goodman, N. 2015.
\newblock Learning the preferences of bounded agents.
\newblock \emph{NIPS Workshop on Bounded Optimality}.

\bibitem[{Evans, Stuhlmueller, and Goodman(2016)}]{evans2016}
Evans, O.; Stuhlmueller, A.; and Goodman, N. 2016.
\newblock Learning the {P}references of {I}gnorant, {I}nconsistent {A}gents.
\newblock In \emph{Proceedings of the Thirtieth AAAI Conference on Artificial
  Intelligence}, 323–329.

\bibitem[{Everitt and Hutter(2018)}]{everitt2018}
Everitt, T.; and Hutter, M. 2018.
\newblock The {A}lignment {P}roblem for {B}ayesian {H}istory-{B}ased
  {R}einforcement {L}earners.
\newblock Technical report.

\bibitem[{Fern et~al.(2014)Fern, Natarajan, Judah, and Tadepalli}]{fern2014}
Fern, A.; Natarajan, S.; Judah, K.; and Tadepalli, P. 2014.
\newblock A {D}ecision-{T}heoretic {M}odel of {A}ssistance.
\newblock \emph{Journal of Artificial Intelligence Research}, 50: 71--104.

\bibitem[{Genewein et~al.(2015)Genewein, Leibfried, Grau-Moya, and
  Braun}]{genewein2015}
Genewein, T.; Leibfried, F.; Grau-Moya, J.; and Braun, D.~A. 2015.
\newblock Bounded {R}ationality, {A}bstraction, and {H}ierarchical
  {D}ecision-{M}aking: {A}n {I}nformation-{T}heoretic {O}ptimality {P}rinciple.
\newblock \emph{Frontiers in Robotics and AI}, 2.

\bibitem[{Gopalan and Mannor(2015)}]{gopalan2015}
Gopalan, A.; and Mannor, S. 2015.
\newblock Thompson {S}ampling for {L}earning {P}arameterized {M}arkov
  {D}ecision {P}rocesses.
\newblock In \emph{Proceedings of The 28th Conference on Learning Theory},
  861--898. PMLR.

\bibitem[{Guez, Silver, and Dayan(2012)}]{guez2012}
Guez, A.; Silver, D.; and Dayan, P. 2012.
\newblock Efficient {B}ayes-{A}daptive {R}einforcement {L}earning using
  {S}ample-{B}ased {S}earch.
\newblock In \emph{Advances in Neural Information Processing Systems},
  volume~25.

\bibitem[{Hadfield-Menell et~al.(2016)Hadfield-Menell, Russell, Abbeel, and
  Dragan}]{hadfield2016}
Hadfield-Menell, D.; Russell, S.~J.; Abbeel, P.; and Dragan, A. 2016.
\newblock Cooperative {I}nverse {R}einforcement {L}earning.
\newblock In \emph{Advances in Neural Information Processing Systems},
  volume~29.

\bibitem[{Ho and Griffiths(2022)}]{ho2022}
Ho, M.~K.; and Griffiths, T.~L. 2022.
\newblock {Cognitive Science as a Source of Forward and Inverse Models of Human
  Decisions for Robotics and Control}.
\newblock \emph{Annual Review of Control, Robotics, and Autonomous Systems}, 5:
  33--53.

\bibitem[{Hu et~al.(2020)Hu, Lerer, Peysakhovich, and Foerster}]{hu2020}
Hu, H.; Lerer, A.; Peysakhovich, A.; and Foerster, J. 2020.
\newblock “{O}ther-{P}lay” for {Z}ero-{S}hot {C}oordination.
\newblock In \emph{International Conference on Machine Learning}, 4399--4410.
  PMLR.

\bibitem[{Javdani, Srinivasa, and Bagnell(2015)}]{javdani2015}
Javdani, S.; Srinivasa, S.~S.; and Bagnell, J.~A. 2015.
\newblock Shared {A}utonomy via {H}indsight {O}ptimization.
\newblock In \emph{Proceedings of Robotics: Science and Systems}.

\bibitem[{Kahneman et~al.(1982)Kahneman, Slovic, Slovic, and
  Tversky}]{kahneman1982}
Kahneman, D.; Slovic, S.~P.; Slovic, P.; and Tversky, A. 1982.
\newblock \emph{{Judgment under uncertainty: Heuristics and biases}}.
\newblock Cambridge university press.

\bibitem[{Knox and Stone(2009)}]{knox2009}
Knox, W.~B.; and Stone, P. 2009.
\newblock Interactively {S}haping {A}gents via {H}uman r{R}inforcement: The
  {TAMER} {F}ramework.
\newblock In \emph{Proceedings of the Fifth International Conference on
  Knowledge Capture}, 9--16.

\bibitem[{Lucas et~al.(2008)Lucas, Griffiths, Xu, and Fawcett}]{lucas2008}
Lucas, C.; Griffiths, T.; Xu, F.; and Fawcett, C. 2008.
\newblock A rational model of preference learning and choice prediction by
  children.
\newblock In \emph{22nd Annual Conference on Neural Information Processing
  System}.

\bibitem[{MacGregor, Ormerod, and Chronicle(2000)}]{macgregor2000}
MacGregor, J.~N.; Ormerod, T.~C.; and Chronicle, E. 2000.
\newblock A model of human performance on the traveling salesperson problem.
\newblock \emph{Memory \& Cognition}, 28(7): 1183--1190.

\bibitem[{Ng and Russell(2000)}]{ng2000}
Ng, A.~Y.; and Russell, S.~J. 2000.
\newblock Algorithms for {I}nverse {R}einforcement {L}earning.
\newblock In \emph{Proceedings of the Seventeenth International Conference on
  Machine Learning}, 663--670.

\bibitem[{Ortega and Braun(2013)}]{ortega2013}
Ortega, P.~A.; and Braun, D.~A. 2013.
\newblock Thermodynamics as a theory of decision-making with
  information-processing costs.
\newblock \emph{Proceedings of the Royal Society A: Mathematical, Physical and
  Engineering Sciences}, 469(2153): 20120683.

\bibitem[{Perez, Such, and Karaletsos(2020)}]{perez2020}
Perez, C.; Such, F.~P.; and Karaletsos, T. 2020.
\newblock Generalized {H}idden {P}arameter {MDPs}: {T}ransferable
  {M}odel-{B}ased {RL} in a {H}andful of {T}rials.
\newblock In \emph{Proceedings of the AAAI Conference on Artificial
  Intelligence}, volume~34, 5403--5411.

\bibitem[{Ramachandran and Amir(2007)}]{ramachandran2007}
Ramachandran, D.; and Amir, E. 2007.
\newblock Bayesian {I}nverse {R}einforcement {L}earning.
\newblock In \emph{IJCAI}, volume~7, 2586--2591.

\bibitem[{Rao(2019)}]{rao2019}
Rao, S.~S. 2019.
\newblock \emph{Engineering {O}ptimization: {T}heory and {P}ractice}.
\newblock John Wiley \& Sons.

\bibitem[{Reddy, Dragan, and Levine(2018)}]{reddy2018}
Reddy, S.; Dragan, A.; and Levine, S. 2018.
\newblock Shared {A}utonomy via {D}eep {R}einforcement {L}earning.
\newblock In \emph{Proceedings of Robotics: Science and Systems}. Pittsburgh,
  Pennsylvania.

\bibitem[{Shah et~al.(2020)Shah, Freire, Alex, Freedman, Krasheninnikov, Chan,
  Dennis, Abbeel, Dragan, and Russell}]{shah2020}
Shah, R.; Freire, P.; Alex, N.; Freedman, R.; Krasheninnikov, D.; Chan, L.;
  Dennis, M.; Abbeel, P.; Dragan, A.; and Russell, S. 2020.
\newblock Benefits of {A}ssistance over {R}eward {L}earning.
\newblock \emph{Workshop on Cooperative AI (Cooperative AI @ NeurIPS 2020)}.

\bibitem[{Shah et~al.(2019)Shah, Gundotra, Abbeel, and Dragan}]{shah2019}
Shah, R.; Gundotra, N.; Abbeel, P.; and Dragan, A. 2019.
\newblock On the feasibility of learning, rather than assuming, human biases
  for reward inference.
\newblock In \emph{International Conference on Machine Learning}, 5670--5679.
  PMLR.

\bibitem[{Sutton and Barto(2018)}]{sutton2018}
Sutton, R.~S.; and Barto, A.~G. 2018.
\newblock \emph{Reinforcement learning: {A}n introduction}.
\newblock MIT press, 2nd edition.

\bibitem[{Viappiani and Boutilier(2010)}]{viappiani2010}
Viappiani, P.; and Boutilier, C. 2010.
\newblock Optimal {B}ayesian {R}ecommendation {S}ets and {M}yopically {O}ptimal
  {C}hoice {Q}uery {S}ets.
\newblock In \emph{Proceedings of the 23rd International Conference on Neural
  Information Processing Systems}, 2352–2360.

\bibitem[{Warnell et~al.(2018)Warnell, Waytowich, Lawhern, and
  Stone}]{warnell2018}
Warnell, G.; Waytowich, N.; Lawhern, V.; and Stone, P. 2018.
\newblock Deep TAMER: {I}nteractive {A}gent {S}haping in {H}igh-{D}imensional
  {S}tate {S}paces.
\newblock In \emph{Proceedings of the Thirty-Second AAAI Conference on
  Artificial Intelligence}, volume~32, 1545--1553.

\bibitem[{Wirth et~al.(2017)Wirth, Akrour, Neumann, F{\"u}rnkranz
  et~al.}]{wirth2017}
Wirth, C.; Akrour, R.; Neumann, G.; F{\"u}rnkranz, J.; et~al. 2017.
\newblock A survey of {P}reference-{B}ased {R}einforcement {L}earning
  {M}ethods.
\newblock \emph{Journal of Machine Learning Research}, 18(136): 1--46.

\end{thebibliography}


\clearpage
\appendix
\section{The GHPMCP algorithm}
\label{app:GHPMCP}
GHPMCP is based on \emph{Monte-Carlo Tree Search}~(MCTS)~\citep{browne2012}. In every planning iteration GHPMCP samples $\omega$ and $\theta$ from respectively a posterior distribution over reward parameters $p_\omega$ and bias parameters $p_\theta$, and simulates $\mathcal{M}_{\omega,\theta}$ down the tree following an Upper Confidence bound for Trees~(UCT) policy~\citep{browne2012}.

\begin{algorithm}[ht]
\caption{GHPMCP}
\label{alg:advice_planning}
\begin{algorithmic}
\REQUIRE{$\mathrm{plan}(s,p_\omega,p_\theta)$}
    \FOR{$i = 1 \dots n\_iterations$}
        \STATE{$\omega \sim p_\omega$}
        \STATE{$\theta \sim p_\theta$}
        \STATE{$h \gets s$}
        \STATE{$\mathrm{simulate}(h,s,\mathcal{M}_{\omega,\theta},1)$}
    \ENDFOR
    \STATE{$h \gets s$}
    \STATE{\textbf{return} $\argmax_a Q(h,s,a)$}

\REQUIRE{$\mathrm{simulate}(h,s,M_{\omega,\theta},d)$}
    \IF{$N(h,s) = 0$} 
        \FORALL{$a \in \mathcal{A}'$}
            \STATE{$N(h,s,a) \gets 0$}
            \STATE{$Q(h,s,a) \gets 0.0$}
        \ENDFOR
    \ENDIF
    \STATE{$a' \gets \argmax_{a \in \mathcal{A}'} Q(h,s,a) + c \sqrt{\frac{\log N(h,s)}{N(h,s,a)}}$}
    \STATE{$s' \sim \mathcal{T}_{\omega, \theta}(\cdot \mid s, a')$}
    \STATE{$r \gets \mathcal{R}_\omega(s, a', s')$}
    \STATE{$h' \gets ha's'$}
    \STATE{$d' \gets d + 1$}
    \IF{$N(h,s,a) = 0$ \OR $d = max\_depth$}
    \STATE{$q \gets r + \gamma \: \mathrm{Est\_Value}(s',M_{\omega,\theta},d')$}
    \ELSE
    \STATE{$q \gets r + \gamma \: \mathrm{simulate}(h',s',M_{\omega,\theta},d')$}
    \ENDIF
    \STATE{$N(h,s) \gets N(h,s) + 1$}
    \STATE{$N(h,s,a) \gets N(h,s,a) + 1$}
    \STATE{$Q(h,s,a) = Q(h,s,a) + \frac{q - Q(h,s,a)}{N(h,s,a)}$}
    \STATE{\textbf{return} $q$}
\end{algorithmic}
\end{algorithm}

Algorithm~\ref{alg:advice_planning} shows this procedure in detail. The function $\mathrm{plan(\dots)}$ runs a predetermined number of planning iterations with sampled $\mathcal{M}_{\omega,\theta}$ and returns the action with the highest Q-value. It takes as input the posteriors $p_\omega$ and $p_\theta$ which we assume are maintained and updated externally. In our implementation we used a particle filter to maintain these beliefs. $\mathrm{simulate(\dots)}$ implements the UCT policy used to recursively simulate $\mathcal{M}_{\omega,\theta}$ down the tree, up to a maximum depth $max\_depth$. In Algorithm~\ref{alg:advice_planning} we have used the variable $h$,  which represents the path from a node to the root of the tree, to identify tree nodes. Variable $d$ represent the current depth. As simulation progresses down the tree, the state and action node visit counts $N(h,s)$ and $N(h,s,a)$ and Q-values $Q(h,s,a)$ of nodes encountered are updated (last lines of the procedure). Simulation proceeds recursively until a leaf node is reached, either because the maximum depth is reached or because a new node is encountered (second if statement). The values of leaf state nodes are estimated using $\mathrm{Est\_Value(\dots)}$, which may implement a roll-out algorithm or a function approximator.

Prior work has introduced algorithms for formalisms similar to GHP-MDPs, but was not applicable here for various reasons. \citet{perez2020} originally proposed to use model predictive control to plan over GHP-MDPs, but that does not account for action informativeness. Thompson sampling has been used to plan in parameterized MDPs, where only the transition function is parameterized~\citep{gopalan2015}. Though this could be extended to GHP-MDPs, Thompson sampling results in policies with high variance. \citet{fern2014} proposed sparse sampling for a similar problem, but that would not scale with large action spaces $\mathcal{A'}$.

\section{Additional details on the day trip design experiment}
\begin{figure}{t}
\centering
\includegraphics[width=0.7\columnwidth]{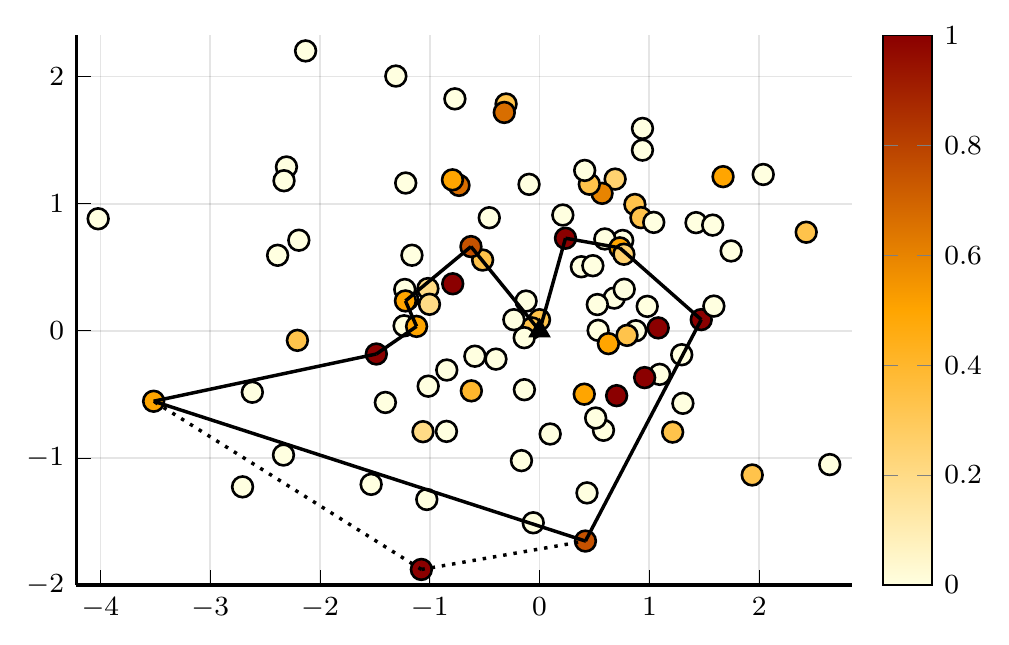}
\caption{In day trip design one looks to choose a set of interesting POIs to visit in a day. This figure shows the locations (axes are coordinates) of a set of available POIs. Their shading correspond to an example agent's interest (lighter color signifies lower interest). An example itinerary is shown, starting at the home location (triangle at the center). The dashed line shows a potential change to the itinerary from additionally visiting the bottom-most POI.}
\label{fig:daytrip}
\end{figure}
\subsection{Objective function}
\label{app:daytrip_design}
We describe here in detail the objective function $f_\omega(s)$ we used in the day trip design experiment. We will first introduce some notation. Let $p_i$ be the $i$-th POI. Any trip $s$ is a subset of the $N = 100$ available POIs $\{p_i\}_{i=1}^{N}$. Let $c(p_i)$ be the admission cost for visiting $p_i$ and let $d(p_i)$ be the time it takes to visit it in minutes. Similarly let $c(s) = \sum_{p_i \in s} c(p_i)$ be the total admission cost of a trip. Finally, let $t_{p_i}^j = 1$ if $p_i$ belong to topic $j$, and 0 otherwise. Let there be $M = 20$ topics. Analogously let $t_a^j = 1$ if the agent is interested in topic j and 0 otherwise. The set of reward parameters $\omega$ then consists of the parameters $\{t_a^j\}_{j=1}^{M}$, and two additional parameters $\mu_c \in [0, \infty)$ and $\sigma_c \in [0, \infty)$ which we will define below.

The objective function $f_\omega$ is a product of two scores: $f_\omega(s) = f^{(1)}_\omega(s) f^{(2)}_\omega(s)$. The first score, $f^{(1)}_\omega(s)$, measures the agent's enjoyment of visiting the POIs in $s$. To define this we must first define an agent's interest, according to $\omega$, in a POI $p_i$. This is defined as the fraction of the POI's topics the agent is interested in:
\[\mathrm{interest}(p_i; \omega) = \frac{\sum_{j = 1}^M t_{p_i}^j t_a^j}{\max \left ( \sum_{j = 1}^M t_{p_i}^j, 1 \right)}\]
The shading of the POIs in Figure~\ref{fig:daytrip} correspond to this function for a random $\omega$. We can now define
\[f^{(1)}_\omega(s) = \frac{\sum_{p_i \in s} d(p_i) \mathrm{interest}(p_i; \omega)}{\mathrm{MAX\_TIME}}\]
where $\mathrm{MAX\_TIME}$ is the maximum allowed duration of a day trip in minutes, $12 \times 60$ in our case.

\begin{figure}[h]
\centering
\includegraphics[width=0.7\columnwidth]{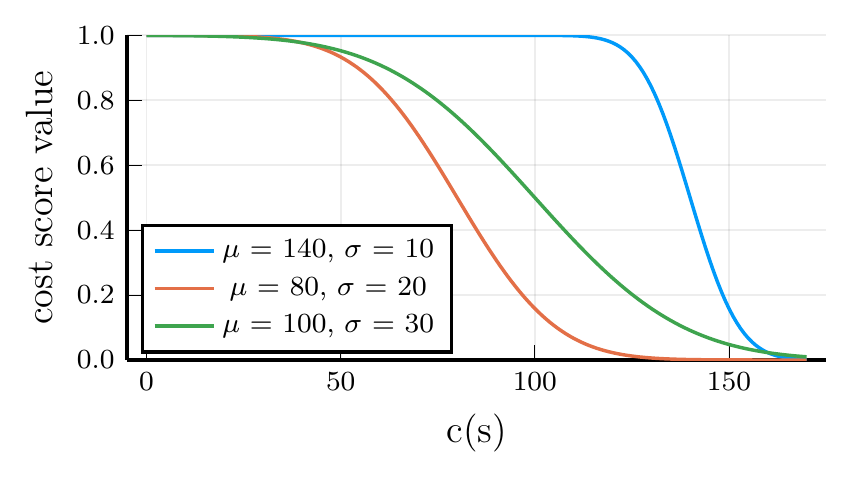}
\caption{Different $f^{(2)}_\omega(s)$ as a function of $c(s)$ for different values for the mean $\mu_c$ and standard deviation $\sigma_c$.}
\label{fig:example_cost_scores}
\end{figure}

The second score, $f^{(2)}_\omega(s)$, determines the agent's willingness to pay the total admission cost of the day trip $c(s)$. It is defined as\footnote{We use the notation $\mathcal{N}_{a}^{b}(\mu, \sigma)$ to denote a normal distribution with mean $\mu$ and standard deviation $\sigma$, truncated on the interval $[a,b]$.}
\[f^{(2)}_\omega(s) = 1 - cdf \left ( \mathcal{N}_{0}^{\infty}(\mu_c, \sigma_c), c(s) \right )\]
where $cdf(d, x)$ is the cumulative density function of distribution $d$ at $x$. Note that for a specific choice of mean $\mu_c$ and standard deviation $\sigma_c$, $f^{(2)}_\omega(s)$ defines the opposite of a sigmoid function. This function has value 1 at $c(s) = 0$ and slopes down to value 0 at $c(s) = \infty$. It crosses the $0.5$ threshold at $\mu_c$ while $\sigma_c$ determines how quickly it slopes down. Figure~\ref{fig:example_cost_scores} visualizes how $\mu_c$ and $\sigma_c$ change $f^{(2)}_\omega(s)$. Intuitively $\mu_c$ roughly determines the agent's cost limit while $\sigma_c$ determines how stringent the agent is with that cost limit.

\subsection{Determining day trip durations} In our implementation we put a soft limit of 12 hours on the length of a day trip. This means that once the duration of an agent's day trip exceeds this, it is not allowed to add further POIs to its trip. The duration of a day trip depends on the time needed to visit the POIs in the day trip ($\sum_{p_i \in s} d(p_i)$ for a day trip $s$) and the time it takes to move between the POIs. To estimate the time it takes to move between two subsequent POIs in an itinerary, we calculate the straight-line distance between them (in kilometers), and divide it by the agent's walking speed. The agent's walking speed is fixed to 5 km/h.

\subsection{Prior distributions and POI generation}
\paragraph{Generating POIs} The POIs used in the experiments are randomly generated. POI attributes are generated from the following priors. The priors were chosen to be representative of a real day trip problem. The x and y coordinates, in kilometers from an artificial center at $(0,0)$, are sampled from a $\mathcal{N}_{-5}^{5}(0, 1.15)$ distribution. The admission cost of a POI is sampled from $\mathcal{N}_{0}^{\infty}(10, 3)$ and the visit time in minutes from $\mathcal{N}_{0}^{100}(30, 20)$. Whether a POI $p_i$ belongs to a topic $j$ is sampled from Bernouilli distribution for every one of the 20 topics: $t_{p_i}^{j} \sim Bern(0.1)$.

\paragraph{Priors over $\Theta$ and $\Omega$} The agents we simulate in our experiments are instantiated with parameter values sampled from the priors $p_{0,\omega}$ and  $p_{0,\theta}$. All parameters have independent prior distributions, which are listed in table~\ref{tab:daytrip_parameters}. The prior distributions were chosen to be realistic for a day trip. The prior over the temperature parameter $\beta_1$ was tuned to yield agents that were optimal enough to benefit from assistance (of any kind, not specifically our method), but not so optimal that they would not need assistance. These priors are also used in the beliefs maintained by the agent, with one minor difference. Within the agent's beliefs, $\beta_1$, the temperature parameter in equation~\ref{eq:own_action_choice}, is always assumed to be 2. This simplifies belief maintenance slightly and had no effect on the quality of assistance.

\begin{table*}[t]
    \centering
    \begin{tabular}{c r  c  p{7cm}}
        & name & prior distribution & explanation \\
        \midrule
        \multirow{3}{*}{$\Omega$} & $t_a^j$ & $Bern(0.3)$ & indicates whether the agent is interested in topic $j$\\
        & $\mu_c$ & $\mathcal{N}(140, 25)$ & parameter of cost the score\\
        & $\sigma_c$ & 10 & parameter of the cost score\\
        \midrule
        \multirow{3}{*}{$\Theta$} & $\beta_1$ & $Unif(1, 4)$ & temperature parameters of the choice in eq.~\eqref{eq:own_action_choice} \\
        & $\beta_2$ & $10 \beta_1$ & temperature parameters of the choice in eq.~\eqref{eq:assistance_choice} \\
        & anchoring bias & $Bern(0.5)$ & if applicable, indicates if the agent has an anchoring bias 
    \end{tabular}
    \caption{Prior distributions for the parameters of the agent model in the day trip design experiment. The left-most column indicates whether the parameter is part of the reward parameters $\Omega$ or the bias parameters $\Theta$}
    \label{tab:daytrip_parameters}
\end{table*}

\subsection{Algorithms and hyperparameters}
\paragraph{Q-Value estimation in the agent model} The Q-values used in the agent model are obtained from a decision tree. This tree is constructed with depth-limited best first search~(BFS) over the agent's view of the problem $\hat{E}$. We ran 500 iterations of BFS with a depth limit of 3.

\paragraph{Planning for assistance} We used GHPMCP for planning over the AIAD formulation to find a policy for the assistant. Table~\ref{tab:planning_parameters} describes the parameter values we used for GHPMCP. Apart from the discounting rate, these parameters were tuned through experimentation. Because the partial automation baseline was implemented using our proposed framework, we used the same hyperparameters as AIAD.

Because automation involves operating directly in the agent's decision problem, an automating policy must be optimal within $E$. But, as the agent's true reward function $R_\omega$ is not known, we can only optimize this policy w.r.t. our current beliefs regarding the reward parameters. To do this we defined a new MDP with stochastic rewards where the reward distribution is obtained from the posterior over the reward parameters. The automation policy, the policy used when automating, was then the policy obtained from planning using a vanilla MCTS implementation on this new MDP. Table~\ref{tab:planning_parameters} describes the parameter values we used. The values themselves were chosen based on experimentation.

\begin{table*}[t]
    \centering
    \begin{tabular}{r  c  c  l}
         & \multicolumn{2}{c}{parameter value} & \\
        \cmidrule{2-3}
        parameter & AIAD & automation & description \\
        \midrule
        $\gamma$ & 0.95 & 0.99 & discounting rate used while planning \\
        $max\_depth$ & 2 & 3 & maximum depth of the tree \\
        $n\_iterations$ & 750,000 & 1,000,000 & number of planning iterations \\
        $c$ & 0.1 & 0.1 & exploration constant used by the UCT policy \\
        $\mathrm{Est\_Value(...)}$ & $ = 0$ & $ = 0$ & function estimating the state value of leaf nodes
    \end{tabular}
    \caption{Parameters used in GHPMCP for AIAD and MCTS for automation in the day trip design experiment. The names of the parameters correspond to those used in Algorithm~\ref{alg:advice_planning}.}
    \label{tab:planning_parameters}
\end{table*}

The assistant's beliefs were maintained using a weighted particle filter with 1024 particles in it. The particles were not refreshed. Because our agent model ${\hat{\pi}(a \mid s, a'; \theta, \omega)}$ is expensive to evaluate we used caching to cache some of its computations. To increase the hit rate of the cache while planning with GHPMCP, we sub-sampled the particle filter. The sub-sampled particle filter contained 100 particles, sampled according to their weights, and was re-sampled for every new planning iteration.

\subsection{Additional results}
\begin{figure}[h]
    \centering
    \begin{subfigure}[b]{0.8\columnwidth}
        \includegraphics[width=\textwidth]{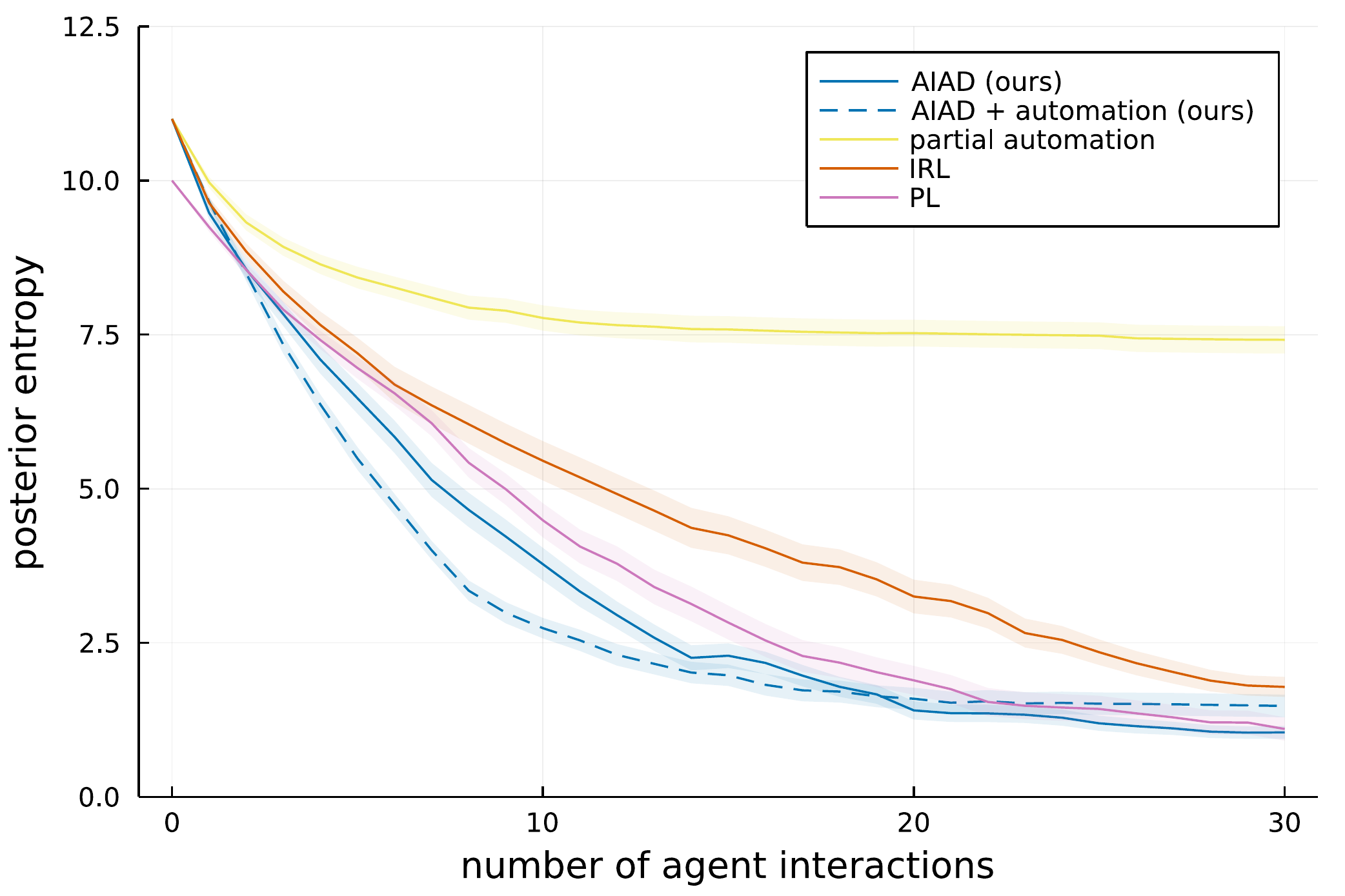}
        \caption{posterior entropy}
        \label{fig:anchoring_experiment_entropy}
    \end{subfigure}
    ~
    \begin{subfigure}[b]{0.8\columnwidth}
        \includegraphics[width=\textwidth]{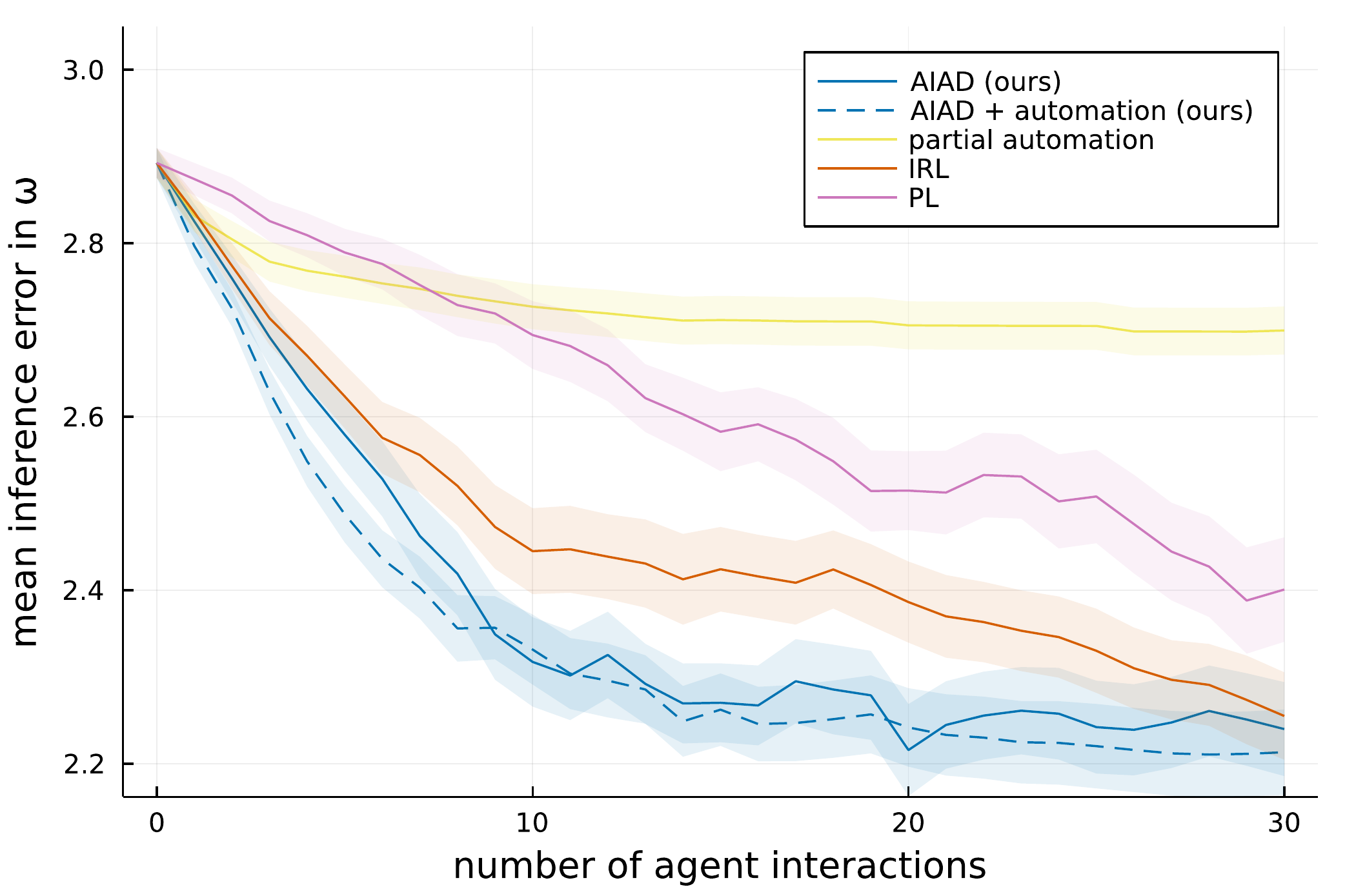}
        \caption{reward parameter estimate error}
        \label{fig:anchoring_experiment_error}
    \end{subfigure}
    \caption{Additional results for the day trip design experiments. \textbf{(a)} Posterior entropy as a function of the number of agent interactions. \textbf{(b)} Mean error between these same methods' posterior over reward parameters and the true reward parameters. The shading shows the standard error of the mean.}
    \label{fig:anchoring_additional_results}
\end{figure}

Here we present additional plots for the day trip design experiment, specifically related to the posterior inferred by the various assistance methods considered. Figure~\ref{fig:anchoring_additional_results} shows the posterior entropy and reward parameter estimation error for every method.

\subsection{Ablation study}
\begin{figure}[t]
    \centering
    \includegraphics[width=0.8\columnwidth]{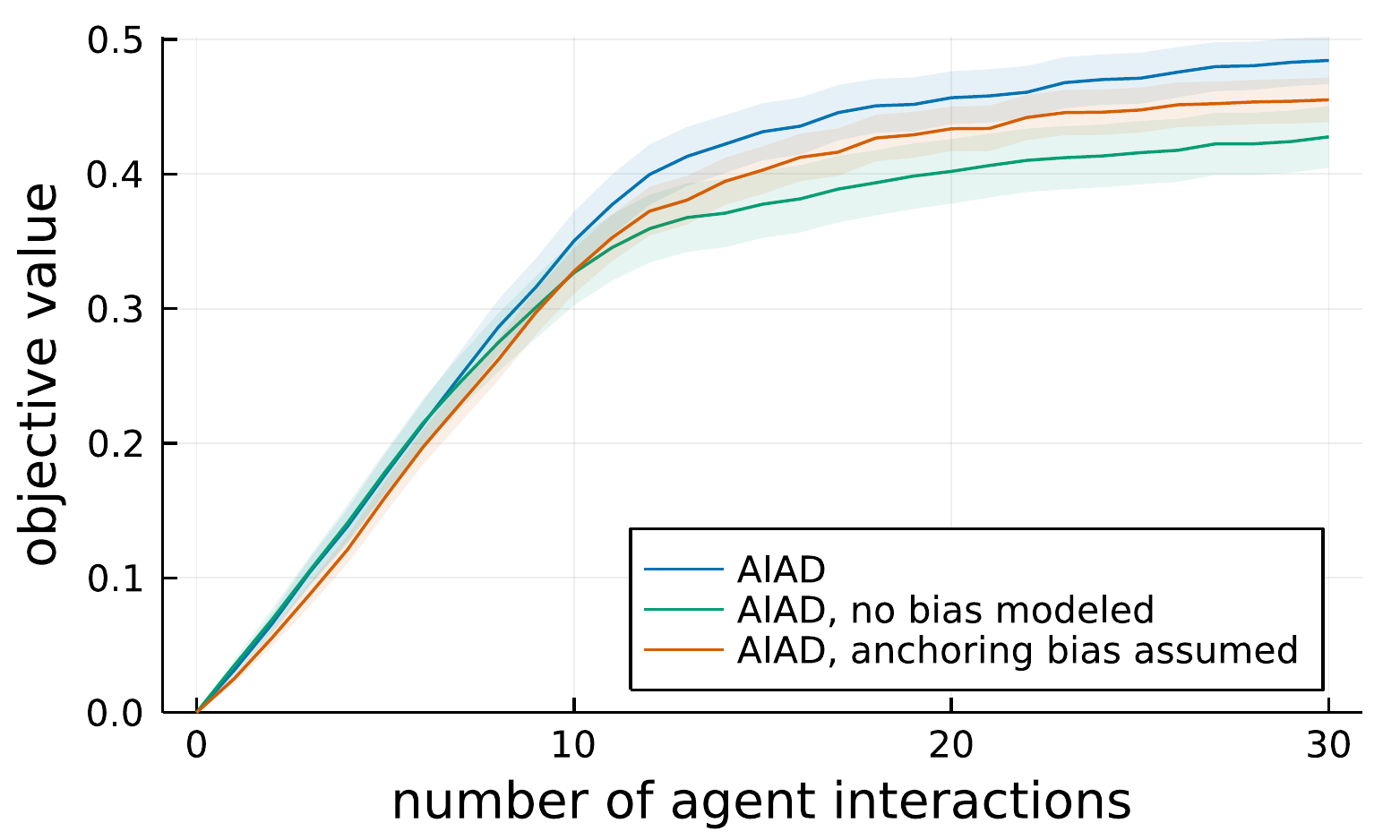}
    \caption{Mean cumulative reward on the day trip design experiment achieved by agents assisted trough AIAD and alternative implementations. The alternatives assume rather than infer the anchoring bias. The shading shows the standard error of the mean.}
    \label{fig:anchoring_ablation}
\end{figure}
To determine how important the online inference of biases is within the proposed AIAD framework, we present here an ablation study in which we consider variants of AIAD which assume rather than infer the anchoring bias in the day trip design experiment. We consider two alternatives: \textbf{AIAD, no bias modeled} does not model the anchoring bias -- or equivalently assumes no agent has it -- and \textbf{AIAD, anchoring bias assumed} assumes that all agents have an anchoring bias and thus always accounts for it. As shown in figure~\ref{fig:anchoring_ablation}, both are significantly outperformed by standard AIAD after 11 interactions. Note that there is initially no benefit in inferring the anchoring bias. Interesting POIs that are close-by are preferred over ones that are far away regardless, so properly accounting for the bias's presence only becomes important once these interesting close-by POIs have been added.

\subsection{Cumulative reward as a function of episode steps}
\begin{figure}[t]
    \centering
    \includegraphics[width=0.8\columnwidth]{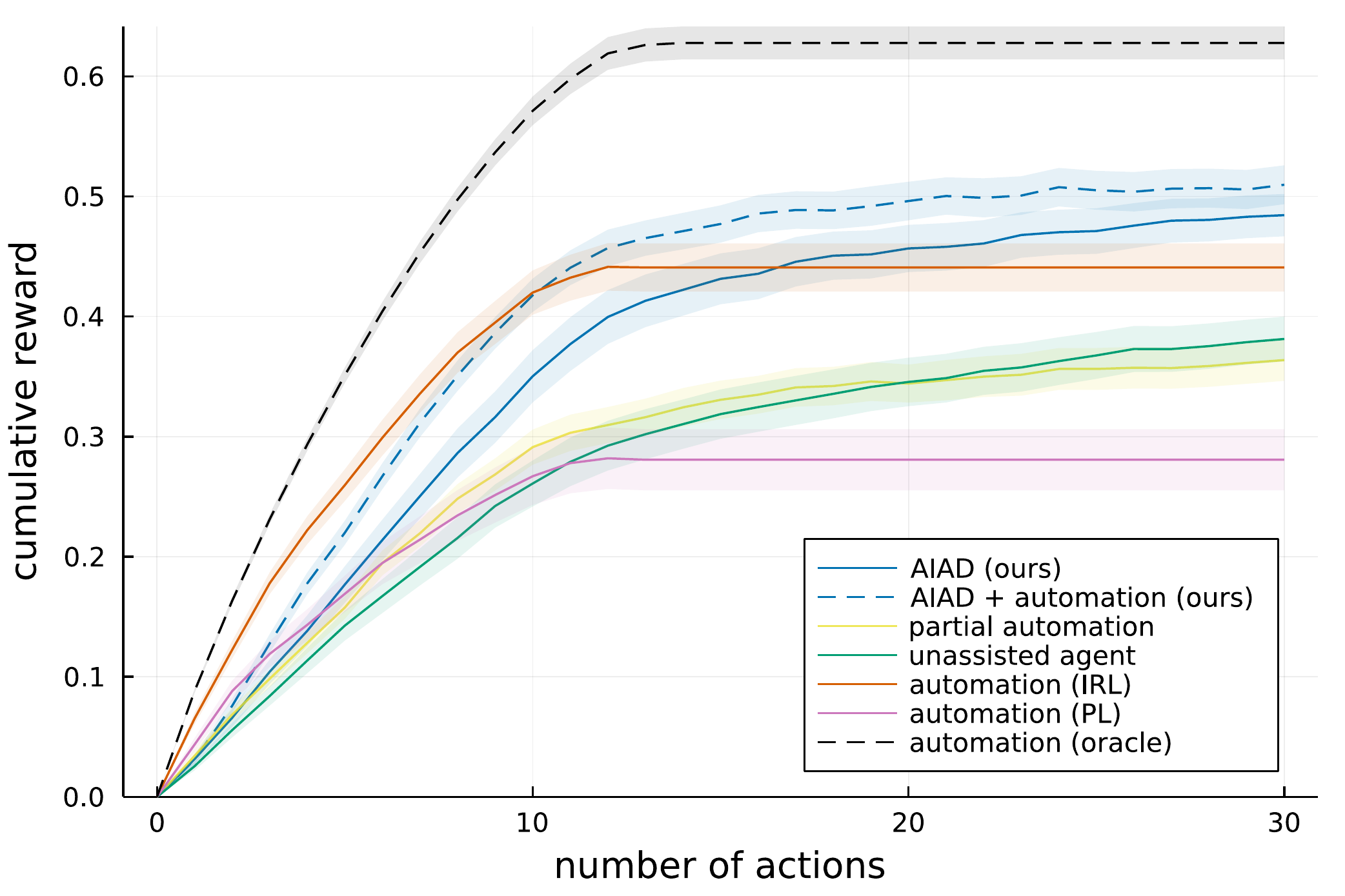}
    \caption{Mean cumulative reward achieved by the different assistance methods considered in this paper as a function of the number of actions taken in the environment for the day trip design experiment. The shading shows the standard error of the mean.}
    \label{fig:anchoring_experiment_cumrew}
\end{figure}

In the main paper we primarily looked at cumulative reward as a function of agent effort. As explained before, for design problems we are most interested in creating the best design possible for a certain amount of effort from the agent. How many actions this takes is not important as long as actions are not taken by the agent. Changing and re-evaluating a design can usually be done very cheaply; the cost of acting in the environment can therefore be ignored.

For completeness we report here on the cumulative reward achieved as a function of the number of actions taken in the environment, regardless of whether the agent, the assistant, or an automating policy took those actions. Figure~\ref{fig:anchoring_experiment_cumrew} show this for the various assistance methods we considered. The IRL and PL baselines are shown here as \textbf{automation (IRL)} and \textbf{automation (PL)} because we only plot the reward obtained using the automating policy. For the sake of comparability to our methods, we ignore here the actions taken by the agent while demonstrating (in the case of IRL) or the queries that happened before automation started (in the case of IRL). Specifically, for these two baselines this graph shows the performance of an automating policy obtained from a posterior inferred on 30 interactions with the agent. Note that this is not an ideal comparison, as it creates an unfair advantage for these baselines. \textbf{automation (oracle)} is the cumulative reward achieved by automation on the true reward function of the agent, and thus represents an upper bound for the other methods.

We observe that an automating policy obtained using IRL initially accumulates reward faster than AIAD and AIAD + automation. However, after about 10 interactions our method starts to outperform the IRL baseline. This is rather impressive as AIAD must learn from observing the agent within the episode, for which it must allow the agent to act in the environment. In contrast, due to how the results are plotted, the automating policies start the episode with all observations already collected.

\section{Day trip design without biases}
\begin{figure}[t]
    \centering
    \includegraphics[width=0.95\columnwidth]{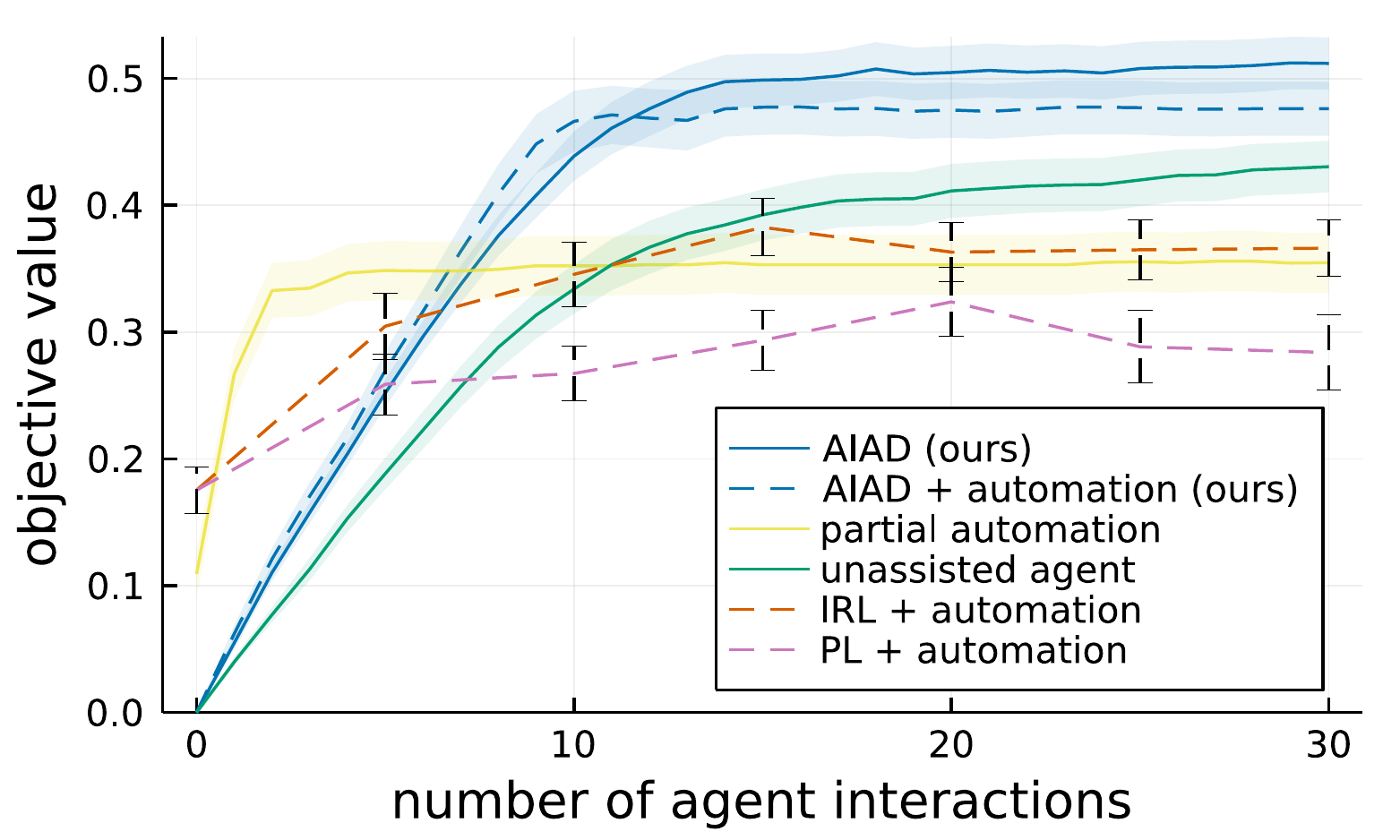}
    \caption{Mean objective value achieved by \emph{known unbiased} agents when assisted by various methods as a function of the number of interactions on the day trip design experiment. The shading and error bars show the standard error of the mean.}
    \label{fig:basic_experiment_performance}
\end{figure}

In the main paper we ran experiments on a day trip design problem in which some agents had an anchoring bias. In this section we describe experimental results in a variant of this problem where no agents have this bias, and contrast the results to the results from the main paper. The assistant knew no biases were present. These experiments were run 50 times with a set-up that was otherwise identical to the original.

Figure~\ref{fig:basic_experiment_performance} shows that AIAD significantly outperforms all baselines after 9 interactions. Comparing these results to the original experiment we see that the anchoring bias reduces the improvement of AIAD over the automation-based baselines. This is expected as the presence of this bias will limit the POIs AIAD can recommend, but does not similarly impact automation. Looking at AIAD + automation, we observe no significant advantage in allowing AIAD to automate. This is not surprising as there are no biases to work around.

\section{Additional details on the inventory management experiment}
We provide additional details and results here for the inventory management experiment introduced in the main paper. In the inventory management problem an agent has to choose in every step some quantity of product to produce for three different products to meet a known stochastic demand. The demand is a Gaussian distribution with known mean and standard deviation.

Our evaluation included the IRL + automation baseline but not the PL + automation baseline. This is because preference learning requires that the agent spends some amount of time before the episode starts to answer the assistant's queries. The need to query before the assistant starts automation is at odds with the online nature of inventory management, and thus we opted not to include the PL + automation baseline. For the case for inverse reinforcement learning, the agent started by taking decision in the environment on its own, and was then eventually replaced by automation. Automation continued from where the agent had stopped.

\subsection{Prior distributions, Algorithms and hyperparameters}
The mean and standard deviation of the demand distributions were sampled independently per time step from a $\mathcal{N}_0^5(2, 0.75)$ and $\chi^2(0.75)$ distribution respectively. The demand was chosen so that the sum of demand of all products would sometimes be higher than the production capacity (which was 12), to ensure that there were times where it was necessary to produce product ahead of high demand. The product profits $v_i$ were independently sampled from a continuous uniform distribution on the range $[0,1]$, though we ensured that one product would always have profit $1.0$. The storage cost $c$, which was the same for all products, was sampled from a $Beta(2.5,8)$ distribution and the future loss of business cost $l$, which was also the same for all products, was sampled from a $Beta(3,3)$ distribution. These values were chosen a priori such that the cost of storage would usually be lower than the cost of lost business and both would easily by offset by the profit from selling product. The bias parameter $\theta$ was sampled from a $\mathcal{N}_{-3}^3(0, 1.5)$ distribution. The width of this distribution was chosen such that the bias would generally cause the agent's demand estimate to shift significantly from the mean of the demand distribution.

\begin{table*}[t]
    \centering
    \begin{tabular}{r  c  c  l}
         & \multicolumn{2}{c}{parameter value} & \\
        \cmidrule{2-3}
        parameter & AIAD & automation & description \\
        \midrule
        $\gamma$ & 0.99 & 0.99 & discounting rate used while planning \\
        $max\_depth$ & 4 & 4 & maximum depth of the tree \\
        $n\_iterations$ & 500,000 & 1,000,000 & number of planning iterations \\
        $c$ & 10.0 & 10.0 & exploration constant used by the UCT policy \\
        $\mathrm{Est\_Value(...)}$ & $ = 0$ & $ = 0$ & function estimating the state value of leaf nodes
    \end{tabular}
    \caption{Parameters used in GHPMCP for AIAD and MCTS for automation in the inventory management experiment. The names of the parameters correspond to those used in Algorithm~\ref{alg:advice_planning}.}
    \label{tab:inventory_planning_parameters}
\end{table*}

\paragraph{Q-Value estimation in the agent model} The Q-values used in the agent model are obtained from a decision tree. This tree is constructed with depth-limited best first search over the agent's view of the problem $\hat{E}$. We ran 300 iterations of BFS with a depth limit of 2. Note that because the agent's point estimates of future demand are floating point values, product inventory within $\hat{E}$ is also represented by floating point values. This is different from the real problem $E$, where the product inventory is always a positive integer number, but where transitions are stochastic.

\paragraph{Planning for assistance} We used GHPMCP for planning over the AIAD formulation to find a policy for the assistant. For the automating policy we used vanilla MCTS as described in the section on additional details about the day trip design experiment above. The parameter values for GHPMCP and MCTS were chosen experimentally, starting from the parameters used in the day trip design experiment. The partial automation baseline used the same hyperparameters as AIAD. The parameters used are listed in table~\ref{tab:inventory_planning_parameters}

As in the day trip design experiment, we used a weighted particle filter to maintain the assistant's beliefs. The computations of the agent model were cached. The particle filter contained 2048 particles and was sub-sampled to 200 particles for planning. Particles were sampled according to their weights, and were re-sampled for every new planning iteration.

\subsection{Additional results}
\begin{figure}[t]
    \centering
    \begin{subfigure}[b]{0.8\columnwidth}
        \includegraphics[width=\textwidth]{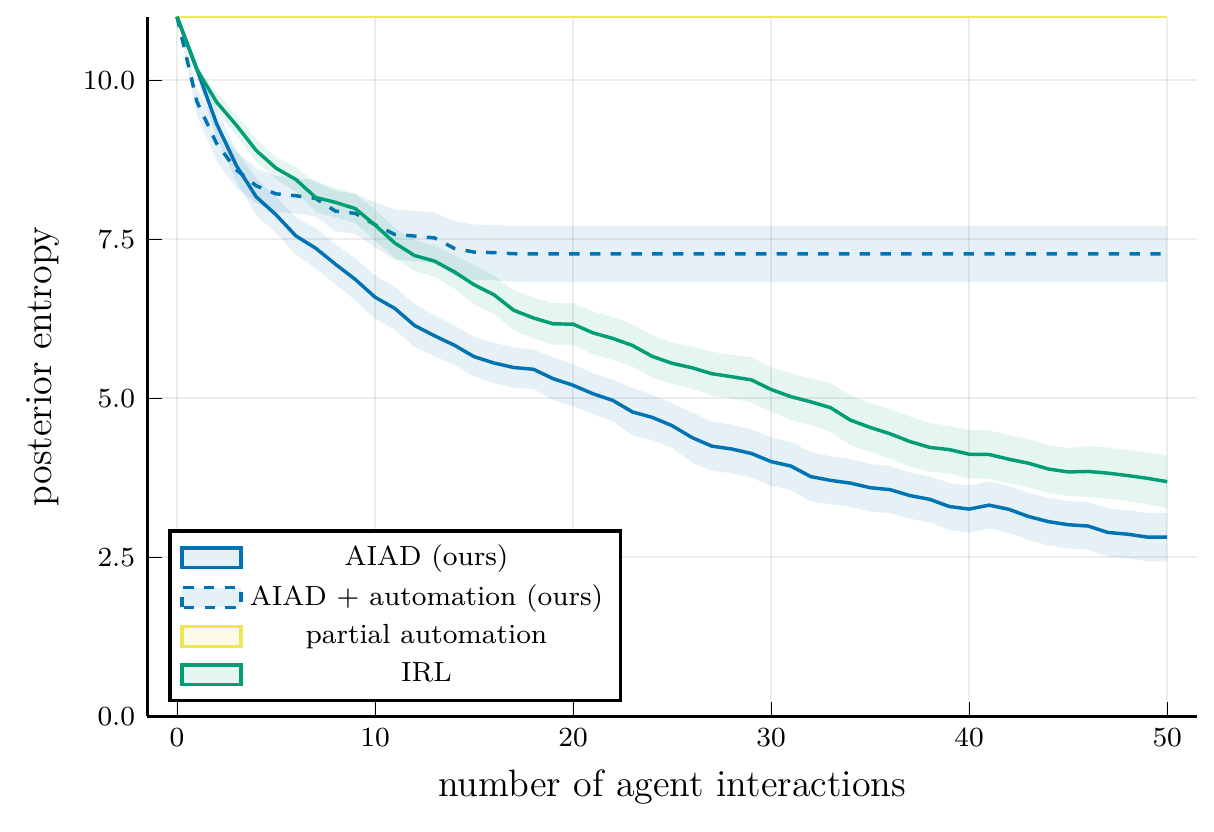}
        \caption{posterior entropy}
        \label{fig:inventory_experiment_entropy}
    \end{subfigure}
    ~
    \begin{subfigure}[b]{0.8\columnwidth}
        \includegraphics[width=\textwidth]{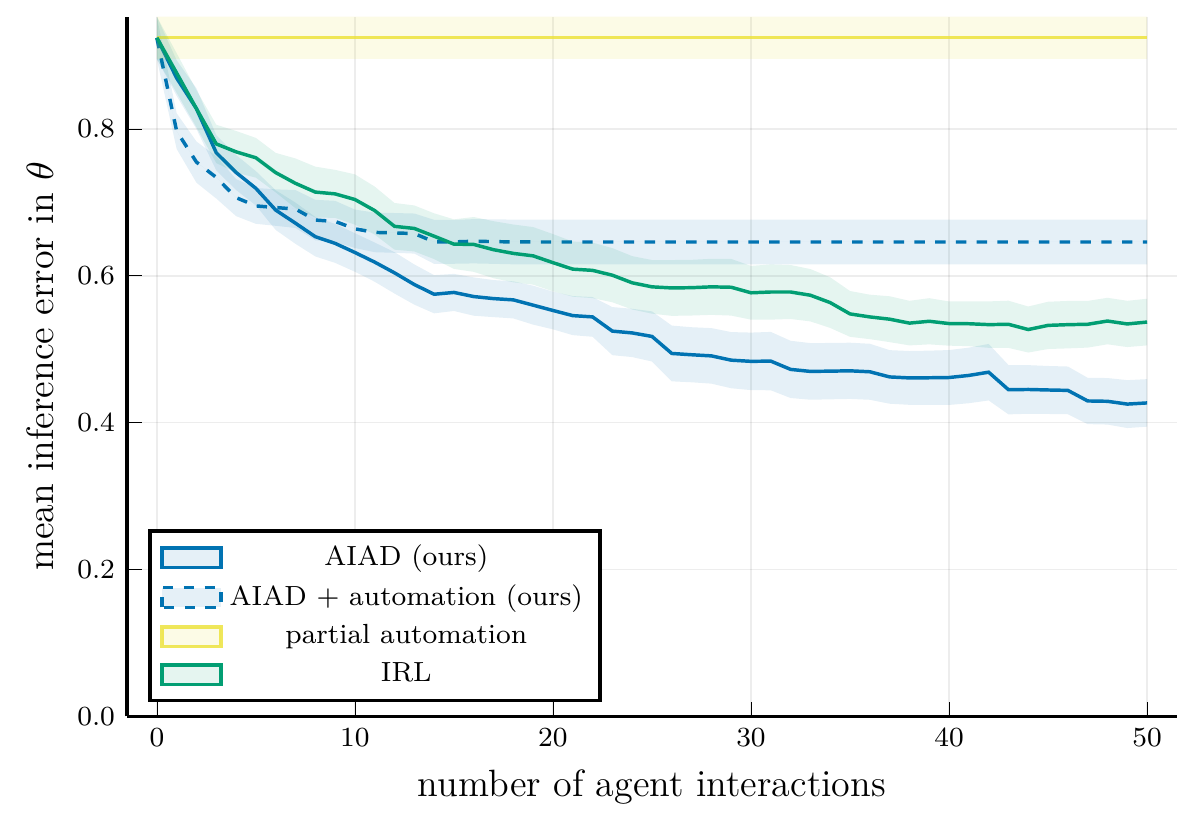}
        \caption{reward parameter estimate error}
        \label{fig:inventory_experiment_error}
    \end{subfigure}
    \caption{Additional results for the inventory management experiments. \textbf{(a)} Posterior entropy as a function of the number of agent interactions. \textbf{(b)} Mean error between these same methods' posterior over reward parameters and the true reward parameters. The shading in both plots shows the standard error of the mean.}
    \label{fig:inventory_additional_results}
\end{figure}

Figure~\ref{fig:inventory_additional_results} shows the posterior entropy and reward parameter estimation error for every method considered in the inventory management problem. Note that it is clear from these plots that for this use case the partial automation baseline does not interact with the agent at all, in contrast to the day trip design experiment where it did.

\subsection{Ablation study}
To study the importance of inferring the agent's optimism/pessimism bias online, we now present an ablation study of AIAD. Here we will compare AIAD to alternative implementations which do not infer the bias but rather assume that it has a certain value. We consider three versions: \textbf{AIAD, no bias assumed} which assumes that $\theta = 0$ and thus that the all agents estimate the average demand correctly, \textbf{AIAD, pessimism assumed} which assumes that all agents have a pessimism bias ($\theta = -1.0$) underestimate the demand, and \textbf{AIAD, optimism assumed} which assumes that all agents have an optimism bias ($\theta = 1.0$) and overestimate the demand. The total cumulative reward over the episode is shown in table~\ref{tab:inventory_management_ablation}. We see that AIAD has a small but statistically significant advantage. 
\begin{table}[ht]
\centering
\begin{tabular}{r c}
method & cumulative reward\\
\midrule
AIAD & $\boldsymbol{185.5 \pm 8.5}$ \\
AIAD, no bias assumed & $175.5 \pm 9.4$ \\
AIAD, pessimism assumed & $173.3 \pm 9.2$ \\
AIAD, optimism assumed & $179.9 \pm 8.8$ \\
\midrule
oracle + automation & $187.6 \pm 7.6$ \\
\end{tabular}
\caption{Mean cumulative discounted reward ($\pm$ standard error) achieved by agents supported by variants of AIAD that assume instead of infer the agents' biases. Bold indicates significantly higher cumulative reward compared to the alternatives.}
\label{tab:inventory_management_ablation}
\end{table}

\section{Does a more accurate user model improve advice?}
\begin{figure}
    \centering
    \includegraphics[width=0.95\columnwidth]{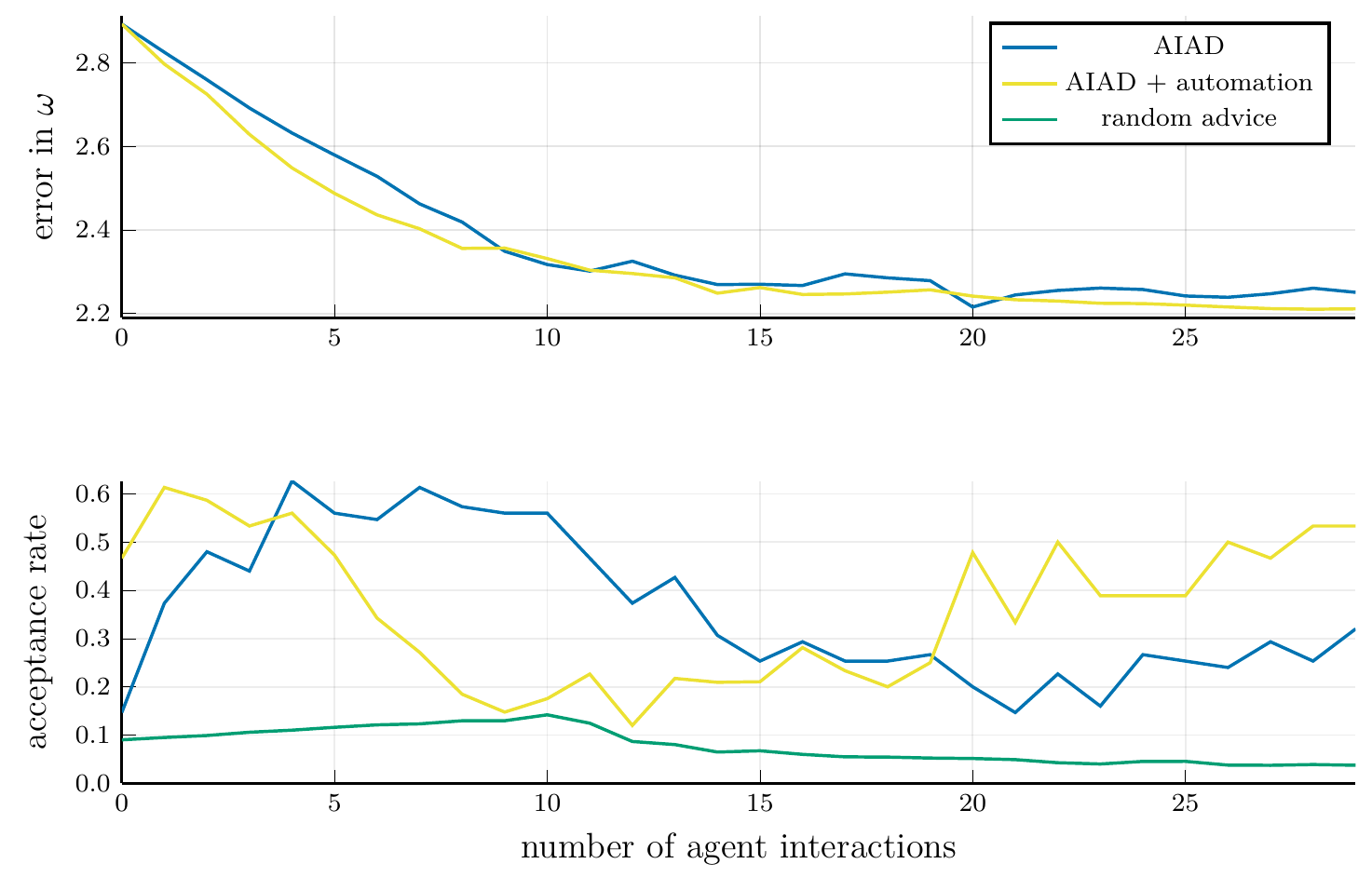}
    \caption{Mean reward inference error and advice acceptance rate of AIAD as a function of the number of interactions with the agent in the day trip design experiment.}
    \label{fig:anchoring_acceptance_analysis}
\end{figure}

To gain a better understanding of AIAD, we would like to know what effect an accurate agent model has on the method. To gain some insight here we will look at the relation between the reward inference error and the acceptance rate of advice provided by AIAD in the day trip design experiment. The reward inference error relates to the accuracy of the agent model, as a higher inference error will lead to less accurate predictions of agent behavior. The acceptance rate relates to the quality of advice. We know that our agent model will reject poor advice, so we know that a low acceptance rate implies that the advice the agent is receiving is bad. Figure~\ref{fig:anchoring_acceptance_analysis} shows the reward inference error and acceptance rate of advice over time for AIAD and AIAD + automation.

For standard AIAD we see a large increase in acceptance rate over the first five interactions with the agent. At the same time we observe a drop in inference error over the same initial period. The improvement in acceptance rate is therefore likely explained by a more accurate user model due to lower error in the reward parameters and more correct inference of the biases. AIAD must first infer the presence of the anchoring bias, meaning that some early recommendations will unavoidably be rejected due to this bias. Analysis of AIAD's inferences shows that it infers $\theta$ with 95\% accuracy after two interactions, which explains the significant increase in acceptance rate over the first two interactions.

From 5 to 10 interactions we see that the acceptance rate remains stable while the reward error continues to decline. But after 10 interactions we observe a strong decline in acceptance rate, not explained by the reward error which at this point has stabilized. A likely explanation is that after 10 interactions the number of acceptable recommendations starts to decline, making it harder to achieve the same acceptance rate. As the design process goes on, and as more POIs are added to the trip, fewer POIs remain that would improve the value of the trip. Once the trip has reached its 12 hour duration limit, POIs can only be removed, which is seldom an interesting prospect for the agent. Therefore, we should expect to see a drop in the number of recommendations the agent is willing to accept. To see where this happens, we have plotted the acceptance rate of random recommendations along the trajectories we observed from AIAD. As expected we see that the acceptance rate drops between 10 and 15 interactions, which can only be explained by a drop in the number of acceptable recommendations. After 15 interactions both the reward error and acceptance rate remain stable.

Comparing AIAD + automation to standard AIAD, we see that it achieves a significantly higher acceptance rate at zero interactions when assisting agents with biases. As no observation have been made yet, the inference error of both of these methods at this point is exactly the same. Looking at figure~\ref{fig:anchoring_experiment_performance}, we know from the fact that the design value at 0 interactions is nonzero that the assistant makes direct edits to the design before it makes the first recommendation to the agent. This means that when the first recommendation is made, the trip is larger than it would be without automation, and thus more POIs are within 500 meters of the itinerary. This makes rejection due to the anchoring bias easier to avoid.

\section{Implementation and computational resources}
Our experiments were implemented in Julia. The code is available under MIT licence from \url{https://github.com/AaltoPML/
Zero-Shot-Assistance-in-Sequential-Decision-Problems}. The implementation relies heavily on the excellent POMDPs.jl ecosystem~\footnote{Maxim Egorov, Zachary N. Sunberg, Edward Balaban, Tim A. Wheeler, Jayesh K. Gupta, and Mykel J. Kochenderfer. POMDPs.jl: A framework for Sequential Decision Making under Uncertainty. Journal of Machine Learning Research, 18(26):1–5, 2017. URL http://jmlr.org/papers/v18/16-300.html} (available under MIT licence). The plots in this paper were generated using Plots.jl.~\footnote{Tom Breloff. Plots.jl. URL https://doi.org/10.5281/zenodo.6523361} A complete list of the packages used is available in the "Project.toml" file that is part of the codebase.

\begin{table*}
    \centering
    \begin{tabular}{ r  c c  c  c  c}
        & \multicolumn{3}{c}{average per run} & \multicolumn{2}{c}{total} \\
        \cmidrule(r){2-4} \cmidrule(r){5-6}
        & memory (GiB) & CPU time & CI (g $\text{CO}_2$)  & CPU time & CI (g $\text{CO}_2$) \\
        \midrule
        AIAD & 3.56 & 10h 9m 33s & 5.49 & 761h 56m 15s & 411 \\ 
        AIAD, no bias & 3.73 & 10h 25m 47s & 5.63 & 825h 13m 45s & 422 \\ 
        AIAD, bias ass. & 2.82 & 6h 46m 26s & 3.66 & 508h 2m 30s & 274 \\ 
        AIAD + automation & 4.32 & 13h 46m 11s & 7.44 & 1032h 43m 45s & 558 \\ 
        partial automation & 1.78 & 3h 19m 09s & 1.79 & 248h 56m 15s & 134 \\ 
        IRL -- inference & 1.16 & 6h 47m 4s & 3.66 & 508h 50m 0s & 274 \\ 
        IRL -- automation  & 1.07 & 37m 23s & 0.34 & 327h 6m 15s & 177 \\ 
        PL -- inference & 0.81 & 2h 31m 5s & 1.36 & 188h 51m 15s & 102 \\ 
        PL -- automation & 1.07 & 22m 35s & 0.20 & 169h 22m 30s & 91 \\ 
        oracle + automation & 1.54 & 1h 16m 55s & 0.69 & 96h 8m 45s & 52 \\ 
        \cmidrule(r){5-6}
         & & & total: & 4624h 11m 15s & 2497 \\ 
    \end{tabular}
    \caption{Computational resources used and associated carbon intensity~(CI) for different assistance methods used in the day trip design experiment.}
    \label{tab:daytrip_resources_anchoring_bias}
\end{table*}

\begin{table*}
    \centering
    \begin{tabular}{ r  c c  c  c  c}
        & \multicolumn{3}{c}{average per run} & \multicolumn{2}{c}{total} \\
        \cmidrule(r){2-4} \cmidrule(r){5-6}
        & memory (GiB) & CPU time & CI (g $\text{CO}_2$)  & CPU time & CI (g $\text{CO}_2$) \\
        \midrule
        AIAD & 3.64 & 11h 35m 48s & 6.26 & 579h 50m 0s & 313 \\ 
        AIAD + automation & 3.66 & 11h 52m 37s & 6.41 & 593h 60m 50s & 321 \\ 
        partial automation & 1.66 & 2h 29m 38s & 1.35 & 124h 41m 40s & 67 \\ 
        IRL -- inference & 1.18 & 4h 29m 13s & 2.42 & 224h 20m 50s & 121 \\ 
        IRL -- automation & 1.02 & 23m 34s & 0.21 & 137h 28m 20s & 74 \\ 
        PL -- inference & 0.81 & 2h 31m 1s & 1.36 & 125h 50m 50s & 68 \\ 
        PL -- automation & 1.49 & 21m 48s & 0.20 & 109h 00m 0s & 59 \\ 
        oracle + automation & 1.5 & 1h 26m 3s & 0.77 & 71h 42m 30s & 39 \\ 
        \cmidrule(r){5-6}
         & & & total: & 1966h 45m 00s & 1062 \\ 
    \end{tabular}
    \caption{Computational resources used and associated carbon intensity~(CI) for various assistance methods considered for the day trip design experiment without biases.}
    \label{tab:daytrip_resources_no_bias}
\end{table*}

\begin{table*}
    \centering
    \begin{tabular}{ r  c c  c  c  c}
        & \multicolumn{3}{c}{average per run} & \multicolumn{2}{c}{total} \\
        \cmidrule(r){2-4} \cmidrule(r){5-6}
        & memory (GiB) & CPU time & CI (g $\text{CO}_2$)  & CPU time & CI (g $\text{CO}_2$) \\
        \midrule
        AIAD & 4.32 & 3h 12m 03s & 1.73 & 64h 1m 0s & 34.57 \\ 
        AIAD + automation & 4.62 & 6h 36m 29s & 3.57 & 132h 9m 40s & 71.4 \\ 
        AIAD, no bias & 4.37 & 2h 53m 49s & 1.56 & 57h 56m 20s & 31.3 \\ 
        AIAD, opt. ass. & 4.52 & 3h 20m 16s & 1.80 & 66h 45m 20s & 36.0 \\ 
        AIAD, pess. ass. & 4.37 & 2h 53m 49s & 1.56 & 57h 56m 20s & 31.3 \\ 
        partial automation & 1.85 & 42m 47s & 0.39 & 14h 15m 20s & 7.7 \\ 
        IRL -- inference & 0.56 & 26m 59s & 0.24 & 8h 59m 40s & 4.9 \\ 
        IRL -- automation & 0.96 & 42m 31s & 0.38 & 70h 51m 40s & 38.3 \\ 
        oracle + automation & 1.7 & 29m 20s & 0.26 & 9h 46m 40s & 5.3 \\ 
        \cmidrule(r){5-6}
         & & & total: & 481h 42m 20s & 260 \\ 
    \end{tabular}
    \caption{Computational resources used and associated carbon intensity~(CI) for various assistance methods considered for the inventory management experiment.}
    \label{tab:inventory_resources}
\end{table*}

Our experiments were run on a private cluster consisting of a mixture of Intel~$^\circledR$ Xeon~$^\circledR$ Gold 6248, Xeon~$^\circledR$ Gold 6148, Xeon~$^\circledR$ E5-2690 v3 and Xeon~$^\circledR$ E5-2680 v3 processors. 

We report in table~\ref{tab:daytrip_resources_anchoring_bias} and table~\ref{tab:daytrip_resources_no_bias} the computational resources used in our day trip design experiments, and in table~\ref{tab:inventory_resources} the computational resources used in our inventory management experiment. As automation was run multiple times per run in the PL and IRL baselines -- after various amounts of interaction with the agent -- we have split reward inference and automation in these tables. The per run figures reported for automation for these baselines is averaged over all times when automation was ran.

The tables additionally contain carbon intensity estimates. For these estimates we have assumed that all experiments were run on an Intel~$^\circledR$ Xeon~$^\circledR$ Gold 6248 CPU running at maximum power. Most of our experiments were run on this specific model of CPU. Our cluster is located in Finland. According to the latest available data (year 2020), the average carbon intensity of one KWh of electricity consumed in Finland is 72 grams.~\footnote{Finngrid Oyj. Real-time CO2 emissions estimate. URL https://www.fingrid.fi/en/484
electricity-market-information/real-time-co2-emissions-estimate/} This yields estimated carbon emissions of $0.00015$ grams $CO_2$ per second of CPU time.

\end{document}